\newcommand{\papermode}{1}

\ifnum\papermode=0
\documentclass[sn-nature,Numbered,iicol]{sn-jnl2}

\else
\documentclass{article}
\usepackage{arxiv}
\usepackage{authblk}
\newcommand{\fnm}[1]{#1} 
\newcommand{\sur}[1]{#1} 
\newcommand{\makeabstract}[1]{%
    \begin{abstract}#1\end{abstract}%
  }
\usepackage{biblatex}
\fi

\usepackage{graphicx}%
\usepackage{multirow}%
\usepackage{amsmath,amssymb,amsfonts}
\DeclareMathOperator*{\argmax}{arg\,max}
\usepackage{amsthm}%
\usepackage{mathrsfs}%
\usepackage[title]{appendix}%
\usepackage{xcolor}%
\usepackage{textcomp}%
\usepackage{manyfoot}%
\usepackage{booktabs}%
\usepackage{algorithm}%
\usepackage{algorithmicx}%
\usepackage{algpseudocode}%
\usepackage{listings}%
\usepackage{multirow}

\usepackage[utf8]{inputenc} 
\usepackage[T1]{fontenc}    
\usepackage{hyperref}       
\usepackage{url}            
\usepackage{nicefrac}       
\usepackage{microtype}      
\usepackage{lipsum}		
\usepackage{doi}
\usepackage{color,soul}
\usepackage{caption}
\usepackage{booktabs}
\usepackage{placeins}
\usepackage{subcaption}
\usepackage{comment}
\usepackage[printonlyused, nohyperlinks]{acronym}
\usepackage{mleftright}
\usepackage{tabularx}
\usepackage{lmodern}
\usepackage{colortbl}

\lstdefinestyle{mystyle}{
  backgroundcolor=\color{backcolour},
  commentstyle=\color{codegreen},
  keywordstyle=\color{magenta},
  numberstyle=\tiny\color{codegray},
  stringstyle=\color{codepurple},
  basicstyle=\ttfamily\footnotesize,
  breakatwhitespace=false,         
  breaklines=true,                 
  captionpos=b,                    
  keepspaces=true,                 
  numbers=left,                    
  numbersep=5pt,                  
  showspaces=false,                
  showstringspaces=false,
  showtabs=false,                  
  tabsize=2,
  escapeinside={(*@}{@*)}
}

\lstdefinestyle{markdownstyle}{
  basicstyle=\ttfamily\small,
  keywordstyle=\color{black},
  commentstyle=\color{black},
  stringstyle=\color{black},
  breaklines=true,
  columns=fullflexible,
  backgroundcolor=\color{gray!10},
  frame=single,
  framerule=0.5pt,
  numbers=none,
  upquote=true
}

\begin{document}

\raggedbottom

\ifnum\papermode=0
\title{Unlocking an End-to-End Multimodal Pathology Foundation Model with Clinical Dialogue}
\else
\title{PRISM2: Unlocking Multi-Modal General Pathology AI with Clinical Dialogue}
\fi

\author[1]{\fnm{Eugene} \sur{Vorontsov}\textsuperscript{*\textdaggerdbl}}
\author[1]{\fnm{George} \sur{Shaikovski}\textsuperscript{*}}
\author[1]{\fnm{Adam} \sur{Casson}\textsuperscript{*}}
\author[1]{\fnm{Julian} \sur{Viret}\textsuperscript{*}}
\author[2]{\fnm{Eric} \sur{Zimmermann}\textsuperscript{*}}
\author[2]{\fnm{Neil} \sur{Tenenholtz}}
\author[1]{\fnm{Yi Kan} \sur{Wang}}
\author[1]{\fnm{Jan H.}\sur{Bernhard}}
\author[1]{\fnm{Ran A.} \sur{Godrich}}
\author[1]{\fnm{Juan A.} \sur{Retamero}}
\author[3]{\fnm{Jinru} \sur{Shia}}
\author[3]{\fnm{Mithat} \sur{Gonen}}
\author[3]{\fnm{Martin R. } \sur{Weiser}}
\author[4]{\fnm{David S.} \sur{Klimstra}}
\author[1]{\fnm{Razik} \sur{Yousfi}}
\author[2]{\fnm{Nicol\`o} \sur{Fusi}}
\author[1]{\fnm{Thomas J.} \sur{Fuchs}}
\author[2]{\fnm{Kristen} \sur{Severson}\textsuperscript{\textdagger}}
\author[1]{\fnm{Siqi} \sur{Liu}\textsuperscript{\textdagger}}

\affil[1]{Paige, NYC, NY United States}
\affil[2]{Microsoft Research, Cambridge, MA United States}
\affil[3]{Memorial Sloan Kettering Cancer Center, NYC, NY United States}
\affil[4]{University of Yale, New Haven, CT United States}

\ifnum\papermode=1
\maketitle
\fi

\makeabstract{Recent rapid progress in the field of computational pathology has been enabled by foundation models. These models are beginning to move beyond encoding image patches towards whole-slide understanding but their clinical utility remains limited. In this work, we present PRISM2, a multimodal slide-level foundation model trained on data from 700,000 diagnostic specimen-report pairs, the largest vision (2.3 million whole slide images) and language (14M question-answer pairs) histopathology dataset to date. By learning through clinical-dialogue supervision, PRISM2 aligns histomorphologic features with the language of diagnostic reasoning, producing slide-level representations that support both direct diagnostic question-answering and transferable embeddings for downstream tasks. Without additional training, PRISM2 matches or exceeds the cancer-detection performance of clinical-grade products. This is observed without loss of generality on other tasks, where PRISM2 achieves top performance. Finally, using survival prediction as the example, we show that task-specific finetuning with a large dataset can outperform task-specific models, further improving performance. These results demonstrate how language-supervised pretraining provides a scalable, clinically grounded signal for learning generalizable pathology representations, bridging human diagnostic reasoning and foundation-model performance.
}

\date{}
\ifnum\papermode=0
\maketitle
\fi

\def\thefootnote{\textdaggerdbl}\footnotetext{ Corresponding author. eugene.vorontsov AT paige DOT ai\\}\def\thefootnote{\arabic{footnote}}

\def\thefootnote{*}\footnotetext{These authors contributed equally to this work.\\}\def\thefootnote{\arabic{footnote}}

\def\thefootnote{\textdagger}\footnotetext{Equal co-supervision.\\}\def\thefootnote{\arabic{footnote}}


\ifnum\papermode=0
\section{Main}
\else
\section{Introduction}
\fi

\begin{figure*}[t]
    \centering
    \begin{subfigure}[b]{\textwidth}
         \centering
         \includegraphics[width=\textwidth]{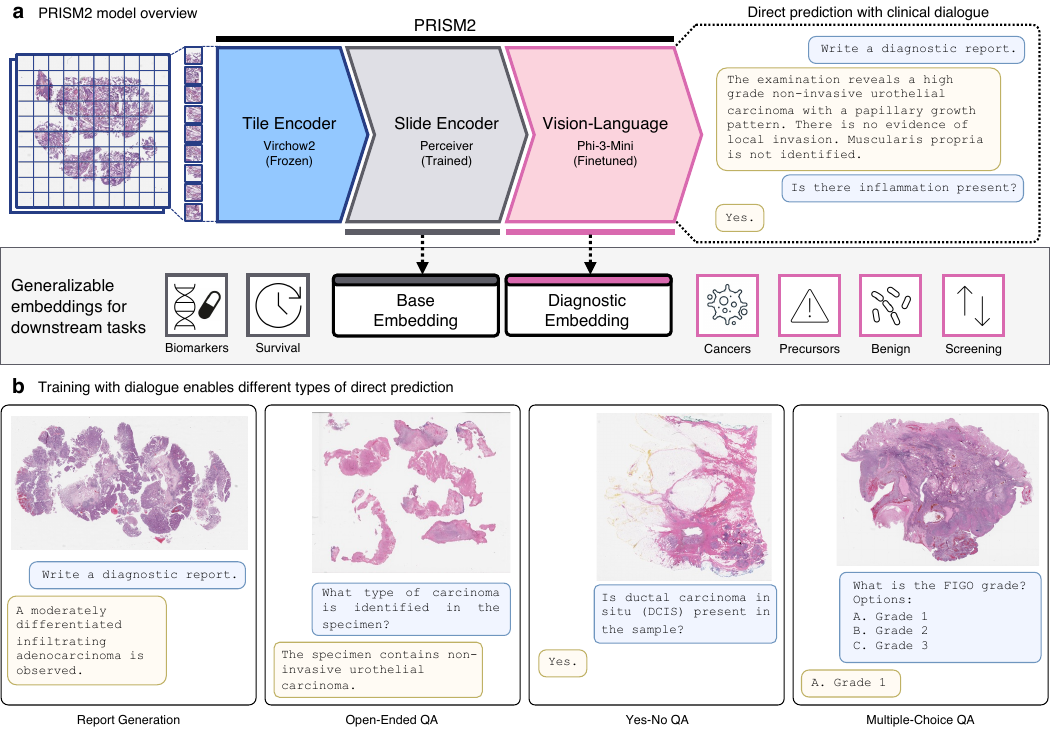}
     \end{subfigure}
    \caption{
    Overview of PRISM2, a slide-level foundation model capable of producing generalizable embeddings as well as direct prediction with clinical dialogue. \textbf{a} The input to PRISM2 is one or more \acp{WSI} associated with a patient-case. These large tissue images are cropped into tiles which are subsequently processed with the Virchow2 foundation model. The sequence of tile embeddings is then aggregated via a slide encoder into a base image representation (``Base Embedding''). The base embedding is then used as input, along with text, to a large vision-language model which produces a diagnostic representation (``Diagnostic Embedding'') and text. The diagnostic embedding is tailored to diagnostic tasks, such as the detection and identification of cancers, precursors to cancers, and benign conditions. Base embeddings are more general and are well suited for transfer learning to tasks outside of the diagnosis-focused training distribution, such as biomarker and survival prediction. \textbf{b} Examples of the four types of direct dialogue-based prediction templates: report generation, open-ended \ac{QA}, yes-no \ac{QA}, and multiple-choice \ac{QA}. Yes-no and multiple-choice \ac{QA} in particular enable directly querying the model for quantifiable predictions with probabilities that can be calibrated.
    }
    \label{fig:teaser}
\end{figure*}

The field of computational pathology has been transformed by the development of foundation models. Models such as Virchow2~\cite{zimmermann2024virchow2}, UNI2~\cite{chen_general-purpose_2023} and H-optimus-1~\cite{h-opt-1} are trained using millions of histopathology tiles spanning diverse tissue types and employing self-supervised objectives~\cite{oquab2023dinov2} to learn generalizable representations. The introduction of foundation models into the computational pathology modeling pipeline has improved performance, robustness, and data efficiency in a variety of tasks such as cancer detection, subtyping, and biomarker quantification \cite{wang2022transformer,chen_general-purpose_2023, vorontsov2024foundation,wang2024screen,campanella2025real}. 
However, a representational gap still exists as most of the desired predictive tasks are at the level of the pathology case, which typically consists of one or more \acp{WSI}, and very few models provide generalizable slide-level embeddings. Instead, current workflows predominantly start from tile representations which are aggregated to create task-specific models via weakly supervised learning. The lack of a slide-level representation results in several shortcomings. For instance, without pretraining, aggregation models are more easily overfit and lack generalizability. They are also likely to require more training data and be less robust to nuisance variation. 

The challenge of creating slide-level foundation models is in choosing a supervisory signal that is useful for compressing the long sequences of tile representations while still being sufficiently generalizable for the large number of possible downstream tasks. Work to date has considered image-only~\cite{xu2024multimodal,chen2022scaling}, paired molecular~\cite{jaume2024transcriptomics,jaume2024multistain,vaidya2025molecular}, and clinical reports~\cite{ahmed2024pathalign,shaikovski2024prism,ding2024multimodal,wang2024pathology} as sources for such signal. Given the consistent observation of the value of scale in foundation models for many different fields~\cite{kaplan2020scaling,zhai2022scaling}, we prioritize signal that is abundantly available and therefore focus our slide-level model pretraining on clinical reports.

Past approaches leveraging clinical reports have suffered several shortcomings. First, many datasets are small, with insufficient diversity to learn generalizable representations. Second, unprocessed pathology report text itself is not enough to provide sufficient supervision. There is heterogeneity in pathology reporting protocols, and these are not always followed; information density can vary accordingly. Finally, many approaches have focused on contrastive objectives to align images and text which misses the opportunity to build on the capabilities of \acp{LLM}.

\begin{figure*}[p]
    \centering
    \begin{subfigure}[b]{\textwidth}
         \centering
         \includegraphics[width=\textwidth]{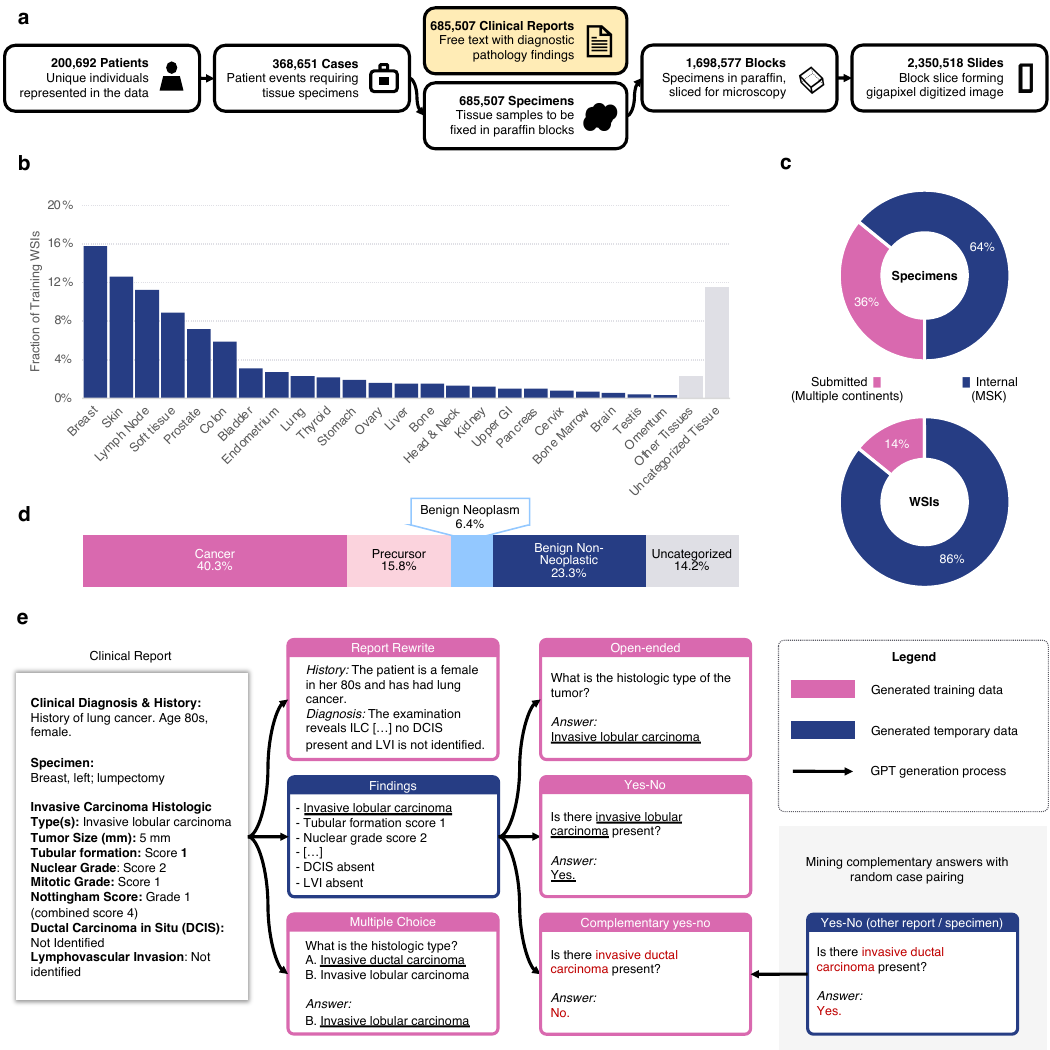}
     \end{subfigure}
    \caption{
    Overview of PRISM2 training data. \textbf{a} The training data can be described in terms of patients, cases, specimens, blocks, and slides, as shown. Clinical reports are paired at the level of specimens. Consequently, PRISM2 is trained at the specimen level to predict diagnostic findings in the clinical reports. \textbf{b} The distribution of tissues represented in the training data. \textbf{c} The distribution of specimens and \acp{WSI} between \ac{MSK} samples and those submitted for second opinion from diverse external sources around the world. \textbf{d} Distribution of high-level conditions in the training data. \textbf{e} Overview of how dialogue examples are generated from clinical reports to be used during PRISM2 training. Every processing step (arrow) uses GPT-4o. Clinical report rewrites are split into clinical history and pathology diagnosis. Diagnostic summaries are produced without demographic, molecular, and sectioning information. Diagnostic findings are first listed, and then converted into yes-no and open-ended question-answer pairs. As the yes-no question distribution is biased toward the findings mentioned in clinical reports, we mine additional complementary yes-no question-answer pairs by assuming that findings which are not mentioned are negative in the specimen. Multiple choice \ac{QA} is derived directly from the reports, following the \ac{CAP} reporting guidelines.
    }
    \label{fig:data}
\end{figure*}

In this work, we present PRISM2 (Fig.~\ref{fig:teaser}, additional detail Extended Data Fig.~\ref{fig:arch}), a slide-level foundation model capable of clinical-grade predictive performance. PRISM2 is the largest multimodal slide-level pathology foundation model to-date and is trained on the largest corpus of data, comprised of 685 thousand distinct clinical reports, corresponding to 2.3 million \acp{WSI} originating from diverse institutions and encompassing different tissue types (Fig.~\ref{fig:data}a-d and Methods~\ref{sec:method_data}). PRISM2 learns to aggregate the signal from tile embeddings and use it as input to a 4 billion parameter \ac{LLM}. Several types of text data are created from the clinical report to provide sufficient and diverse training signal (see Fig.~\ref{fig:data} and Methods~\ref{sec:method_rewrites} for details). This enables PRISM2 to be used for report generation and question-answering (see Extended Data Fig.~\ref{fig:extended_dialogue} for examples). 
While PRISM2 is trained to replicate clinical dialogue, the goal of this supervision is not to deploy the model as a conversational agent for clinical practice but to use dialogue as a rich supervisory signal that encourages the learning of semantically grounded and transferable slide-level representations.

PRISM2 is trained using a two-stage approach (see Methods~\ref{sec:arch} and~\ref{sec:method_training} for additional details). In the first stage, a perceiver-based~\cite{jaegle2021perceiver} slide encoder, using Virchow2~\cite{zimmermann2024virchow2} tile embeddings as input, is trained via a contrastive objective as well as an auto-regressive objective, similar to CoCa~\cite{yu2022coca,shaikovski2024prism}. Clinical text is used in both objectives: BioGPT~\cite{luo2022biogpt} is used to create embeddings and Phi-3 Mini~\cite{abdin2024phi} is the language model for the contrastive and auto-regressive objectives, respectively. In the second stage, the image encoder is frozen and only the language model is fine-tuned using the auto-regressive objective. The choice of architecture and two-stage training approach yield two types of slide-level embeddings: the \textit{base} embedding, defined as the output of the slide encoder, and the \textit{diagnostic} embedding, defined as the language model's hidden state. Empirically, we observe that the base embedding generalizes better to tasks not typically observed in the clinical report such as biomarker prediction and prognosis. Alternatively, the diagnostic prediction is tuned to tasks such as cancer detection and subtyping. 

PRISM2 is evaluated on a suite of diagnostic classification, biomarker prediction, and survival tasks in several different manners aiming to capture the breadth of slide-level foundation model applications. Direct prediction, i.e. without further training of PRISM2, is performed using question answering. Linear-probe evaluation, i.e. with an additional linear predictive model, is used to measure how well the embeddings generalize to both in- and out-of-distribution tasks. In all cases, evaluation data is held out at the patient level from PRISM2 and Virchow2 training data. Overall, we show improved performance by using the language capabilities, while maintaining state-of-the-art performance in applications not typically represented in the clinical report, e.g. in the case of biomarkers. Notably, we show that, without further training, PRISM2 can achieve clinical-grade performance for cancer detection. We conclude with an assessment of forward looking goals, moving beyond detection and subtyping and demonstrate the first application of a slide-level foundation model for clinical report completion, following College of American Pathologists (\ac{CAP}) reporting guidelines. Taken together, these results demonstrate the value of large-scale, multimodal training.

\section{Results}

\subsection{Clinical dialogue training achieves diagnostic-grade performance without task-specific training}
\label{sec:direct}

\begin{figure*}[t]
    \centering
    \includegraphics[width=0.98\linewidth]{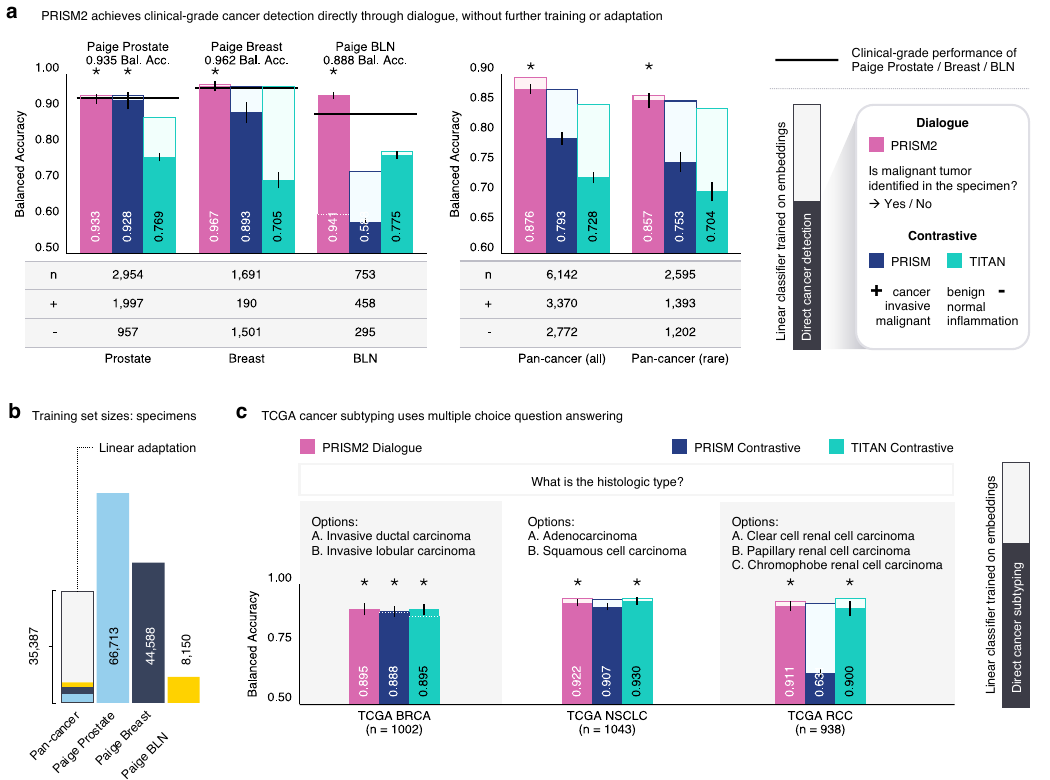}
    \caption{
    Direct prediction through dialogue with PRISM2. \textbf{a} Yes-no \ac{QA} matches or exceeds the performance of existing clinical-grade products for the detection of invasive cancer as compared to Paige Prostate, Paige Breast, and Paige Breast Lymph Node (BLN), when evaluated on the corresponding product testing datasets. Yes-no \ac{QA} on pan-cancer and rare variants exceeds the contrastive performance of competing approaches. After training a linear classifier using the diagnostic embedding as input on a large pan-cancer dataset, PRISM2 pan-cancer performance improves, and continues to outperform competing approaches with similar adaptation. \textbf{b} The size of the pan-cancer training set used for linear adaptation is similar to the training sets of the three clinical models; however, prostate (light blue), breast (dark blue), and BLN (yellow) tissues are a relatively small subset of the pan-cancer dataset. \textbf{c} Direct prediction performs well on \ac{TCGA} cancer subtyping with multiple-choice \ac{QA}. \textit{n} indicates number of samples; $+$ and $-$ indicate number of positive and negative samples, respectively. Error bars show the 95\% confidence interval computed with bootstrapping. * direct prediction result tied for first place (p < 0.05, permutation test).
    }
    \label{fig:direct_prediction_results}
\end{figure*}

Unlike tile-level computational pathology foundation models which are trained using a pretext task, PRISM2 is trained to generate clinically relevant text. As a result, for certain tasks, the model can have sufficiently high performance without further training, a setting we call \textit{direct prediction}. PRISM2 is able to perform direct prediction via its question answering capabilities. To-date, direct prediction in the computational pathology literature has focused on the contrastive setting (e.g.~\cite{ding2024multimodal,shaikovski2024prism}), where candidate labels are compared with image embeddings and the nearest candidate label, measured by distance in embedding space, is selected. Using cancer detection and subtyping tasks for evaluation, we measure the direct prediction of PRISM2 using question-answering and PRISM and TITAN using contrastive comparison (see Sec.~\ref{sec:eval} for further methodological details on the evaluation protocols, Sec.~\ref{sec:comp_models} for further details on the comparison models and Sec.~\ref{sec:eval_tasks} for details on the evaluation datasets).

\begin{figure*}[p]
    \centering
    \includegraphics[width=0.98\linewidth]{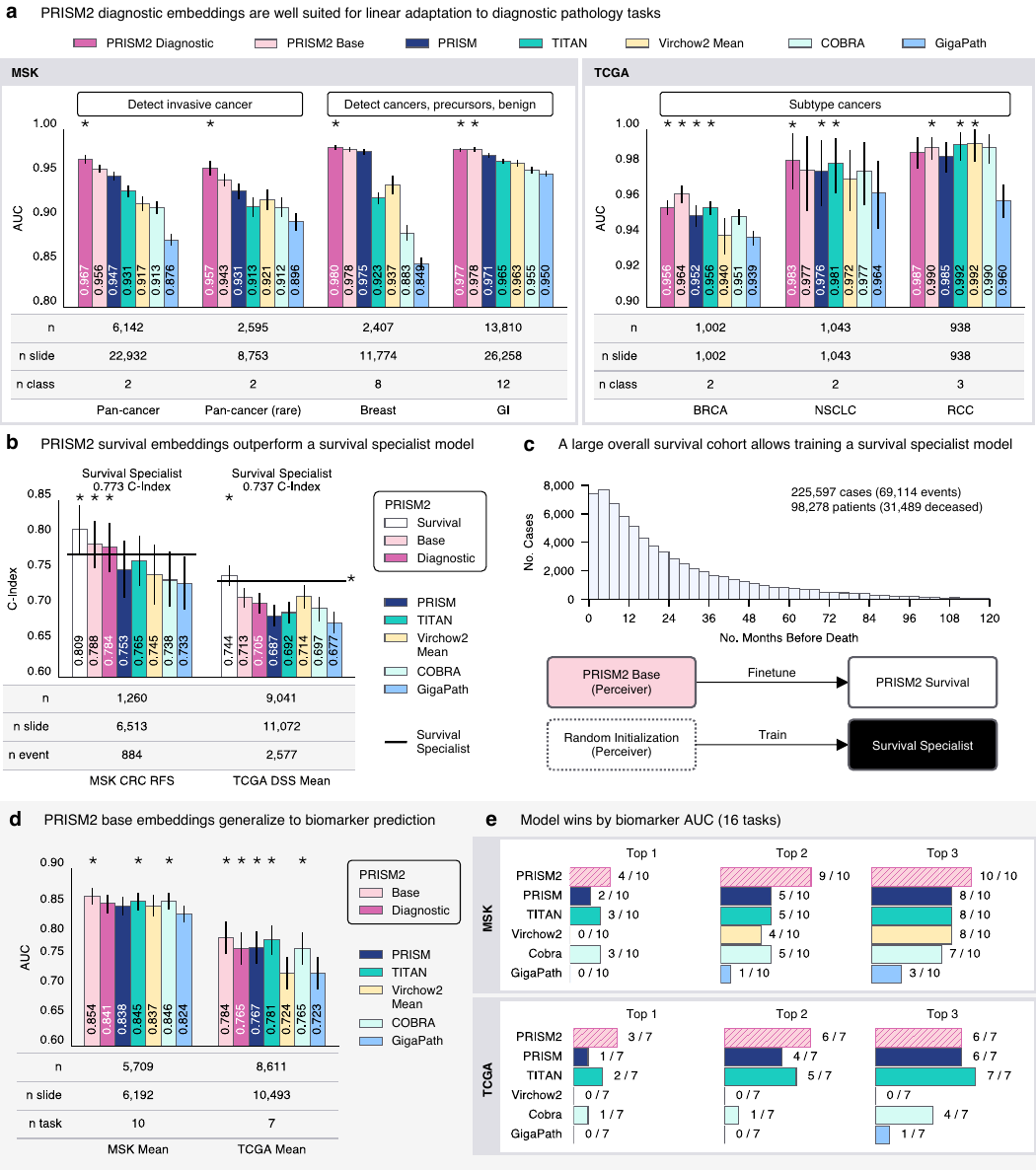}
    \caption{
    Adapting slide-level model embeddings to downstream tasks. \textbf{a} Linear probing with PRISM2 diagnostic embeddings outperforms all other embeddings on a set of challenging tasks including pan-cancer detection and the detection and identification (typing) of cancers, precursors to cancer, and benign conditions in breast and \ac{GI} tissues. Most embeddings perform similarly on \ac{TCGA} cancer subtyping. \textbf{b}-\textbf{e} PRISM2 base embeddings are better suited for generalization beyond diagnostic tasks. \textbf{b} Survival task performance evaluated with Cox regression for \Acf{CRC} \acf{RFS} with \ac{MSK} data and \acf{DSS} averaged over 13 tasks with \ac{TCGA} data. PRISM2 Survival embeddings outperform all others and match or exceed the 
    }
    \label{fig:adaptation_results}
\end{figure*}
\begin{figure*}[!t]
    \ContinuedFloat
    \caption*{(continued) performance of the survival specialist model, followed by PRISM2 Base. \textbf{c} A large \acf{OS} cohort was used to train the survival specialist model as well as the PRISM2 Survival model, by fine-tuning the slide encoder (base embedding). \textbf{d} PRISM2 base embeddings perform at least as well as those of any other model for biomarker detection, as evaluated via linear probing. \textbf{e} The number of times each model scored in the top 1, top 2, and top 3 across the 10 \ac{MSK} and 7 \ac{TCGA} biomarker tasks (measured with \ac{AUC} rounded to two decimal points). \textit{n} indicates number of samples; \textit{n slide}, \textit{n class}, \textit{n event}, and \textit{n task} indicate number of \ac{WSI}s, classes, recurrence or death events, and tasks, respectively. Error bars show the 95\% confidence interval computed with bootstrapping. * linear probe result tied for first place (p < 0.05, permutation test).
    }
    \hrule
\end{figure*}

First, we demonstrate direct prediction performance on tissue-specific cancer detection (Fig.~\ref{fig:direct_prediction_results}a). We compare to specialist commercial models, specifically Paige Prostate, Paige Breast and Paige \ac{BLN}. These models were developed with rigorous clinical validation and large, curated datasets, as summarized in Fig.~\ref{fig:direct_prediction_results}b. Paige Prostate has FDA De Novo approval for clinical use in the US. Paige Breast Lymph Node has an FDA Breakthrough Device designation. All three products are CE-IVD and UKCA marked for clinical diagnostic use in Europe and the UK. Using yes-no question answering (Supplementary Table~\ref{tab:supp_dialogue_yn_prompts}), PRISM2 matches the cancer detection performance of Paige Prostate and Paige Breast, and outperforms Paige \ac{BLN} without further training (see Fig.~\ref{fig:direct_prediction_results}a). For comparison, neither PRISM nor TITAN match the performance of the clinical-grade products with direct (contrastive) prediction and have a particularly notable performance gap on \ac{BLN}. PRISM and TITAN are able to improve with linear adaptation (shown in the shaded region) but still trail PRISM2 performance.

Moving to pan-cancer prediction, again PRISM2 has the best performance overall (Fig~\ref{fig:direct_prediction_results}a). While PRISM2 direct prediction is similar to linear pan-cancer prediction on prostate, breast, and \ac{BLN} product testing datasets, linear adaptation shows some improvement on the pan-cancer test set which includes 9 common and 7 rare cancer types by tumor origin tissue. This suggests that for some cancer types, model adaptation by training with PRISM2 diagnostic embeddings can improve performance over direct prediction. Indeed for all three models, linear adaptation improves performance. Nevertheless, as with direct prediction, only PRISM2 matches or exceeds the performance of all three clinical-grade products with linear adaptation.

We also consider direct subtyping prediction using TCGA data~\cite{weinstein2013cancer,liu2018integrated}, specifically for breast, lung and kidney cancers (Fig~\ref{fig:direct_prediction_results}c; see Supplementary Tab.~\ref{tab:supp_dialogue_mcq_prompts} for multiple-choice question prompts). In these cases, the models largely have the same performance, except PRISM in the kidney task.

To calculate balanced accuracy (Fig~\ref{fig:direct_prediction_results}a), both the direct and linear predictions use a probability threshold at 0.5, whereas, the clinical-grade products each have a carefully tuned prediction threshold. However, the threshold-free \ac{AUC} performance (Extended Data Fig.~\ref{fig:extended_direct_prediction_results}a) yields the same conclusions: PRISM2 direct prediction can match or exceed the performance of clinical-grade products in cancer detection but there exists no probability threshold with which either PRISM or TITAN could do the same, even with linear probing trained on a large-scale pan-cancer dataset.

\subsection{PRISM2 diagnostic embeddings are well suited to diagnostic pathology tasks}

As noted above, as a result of model architecture and algorithmic training design choices, PRISM2 generates two types of embeddings: a base embedding and a diagnostic embedding. As is standard for foundation models, we expect these embeddings to generalize to many downstream tasks. To evaluate these representations, we gathered several datasets to assess different models' ability to identify cancers, precursors, and benign conditions. Specifically, we gather pan-cancer, pan-cancer (rare), breast, and \ac{GI} detection tasks from \ac{MSK}, and BRCA, NSCLC, and RCC subtyping from TCGA~\cite{weinstein2013cancer,liu2018integrated}. The large \ac{MSK} datasets present challenging real-world pathology tasks. The pan-cancer task involves the detection of invasive cancer across 16 cancer tissues of origin (7 of which comprise the ``rare'' cancers subset). The breast and GI tasks involve the detection of cancers, precursors to cancer, and benign conditions (see Extended Data Fig.~\ref{fig:extended_diagnostic_results} for detailed per-condition results). The TCGA tasks involve cancer subtyping in breast, lung, and kidney. Additional dataset details can be found in Sec.~\ref{sec:eval_tasks}.

As comparison slide-level embeddings, we evaluate PRISM~\cite{shaikovski2024prism}, TITAN~\cite{ding2024multimodal}, COBRA~\cite{lenz2025unsupervised}, Prov-GigaPath~\cite{xu2024whole}, and a tile-level baseline using only the mean of the Virchow2~\cite{zimmermann2024virchow2} embeddings. The mean tile embedding is an important baseline as it represents a naive aggregation of the tile embeddings. In all cases, the slide-embedding is used as input to a linear model. For further details regarding the comparison methodology and models, refer to Sec.~\ref{sec:eval} and~\ref{sec:comp_models}, respectively.

Fig.~\ref{fig:adaptation_results}a presents performance on standard clinical tasks.  
In all settings, both PRISM2 base and diagnostic embeddings perform as well as or better than other embeddings. For pan-cancer detection in particular, PRISM2 diagnostic embeddings (0.976 AUC on all cancers) substantially outperform both PRISM2 Base embeddings (0.956 AUC) and all other embeddings (0.947 AUC with PRISM, 0.931 AUC with TITAN). Furthermore, the performance drop on the rare cancers subset is modest (0.967 AUC to 0.957 AUC). In breast, PRISM2 and PRISM embeddings outperform others. In \ac{GI}, we observe a marked performance drop for TITAN on precursors to cancer and benign conditions (Extended Data Fig.~\ref{fig:extended_diagnostic_results}), whereas PRISM2 performance remains high across all conditions. Interestingly, although the COBRA slide-level model uses Virchow2 tile embeddings, its embeddings often under-perform the baseline method of taking a mean of the Virchow2 tile embeddings, suggesting the importance of the slide-level training stage.

\subsection{PRISM2 survival embeddings outperform a survival specialist model}

Accurate prognostic prediction is of major interest for the computational pathology field but generally understood to be very challenging both because of lack of data as well as uncertain connections between morphology and patient outcome~\cite{song2023artificial,shmatko2022artificial}. We assess the benefit of pretraining with report data to survival prediction tasks as well as the value of additional finetuning using a large survival dataset. To do so, we gather a dataset of over 225 thousand cases with overall survival (\ac{OS}) information, across nearly 100 thousand patients (Fig.~\ref{fig:adaptation_results}c). Using this data, we  fine-tune the PRISM2 slide encoder as well as train a survival specialist model from scratch (see Sec.~\ref{sec:method_training} for additional detail). We refer to the embeddings of the finetuned model as \textit{survival} embeddings. We select \ac{CRC} recurrence-free survival \ac{RFS} prediction with \ac{MSK} data and 13 disease-specific survival \ac{DSS} prediction tasks with \ac{TCGA} data for evaluation (Sec.~\ref{sec:eval_tasks} for dataset details). 

The results of the survival analysis are presented in Fig.~\ref{fig:adaptation_results}b. PRISM2 survival embeddings exceed the performance of the survival specialist model, especially for \ac{MSK} \ac{CRC} \ac{RFS} (0.809 vs 0.773 C-Index). Without survival fine-tuning, PRISM2 base embeddings outperform all other model embeddings for \ac{MSK} \ac{CRC} \ac{RFS} and match the Virchow2 Mean baseline (mean of Virchow2 tile embeddings) on \ac{TCGA} \ac{DSS}. Unsurprisingly, the base embeddings slightly outperform the diagnostic embeddings, as survival information was not present in the dialogue training data. See Extended Data Fig.~\ref{fig:extended_survival_results} for detailed per-task results.

\subsection{PRISM2 base embeddings generalize to biomarker prediction}
Biomarker prediction using \ac{HE} \acp{WSI} is another task that, like survival prediction, could extend the applications of computational pathology. Early results have shown great promise in this area~\cite{wang2024screen,campanella2025real}. We select two sets of biomarkers, from \ac{MSK} and \ac{TCGA} to evaluate prediction performance using embeddings (Sec.~\ref{sec:eval_tasks} for dataset details).
Fig.~\ref{fig:adaptation_results}d,e (and Extended Data Fig.~\ref{fig:extended_biomarker_results}) presents the results. Similarly to survival prediction, PRISM2 base embeddings tend to outperform diagnostic embeddings. While the average \acp{AUC} on \ac{MSK} \ac{TCGA} tasks are the highest for PRISM2 base embeddings, as compared to diagnostic embeddings, PRISM, TITAN, Virchow2 Mean (mean of the tile embeddings), COBRA, and GigaPath, the next-best performing models are close: 0.846 (vs 0.854) AUC with COBRA on \ac{MSK} and 0.781 (vs 0.784) AUC on \ac{TCGA} with TITAN. Similarly, looking at Fig.~\ref{fig:adaptation_results}e, PRISM2 embeddings (either base or diagnostic) score the most often among the top 1, top 2, and top 3 best-performing models per biomarker prediction task but only by a small margin (Fig.~\ref{fig:adaptation_results}e). Overall, we can conclude that PRISM2 embeddings are, on average, at least as good as those of other models for biomarker prediction.

\subsection{Dialogue unlocks pathology report completion}

\begin{figure*}[p]
    \centering
    \includegraphics[width=0.98\linewidth]{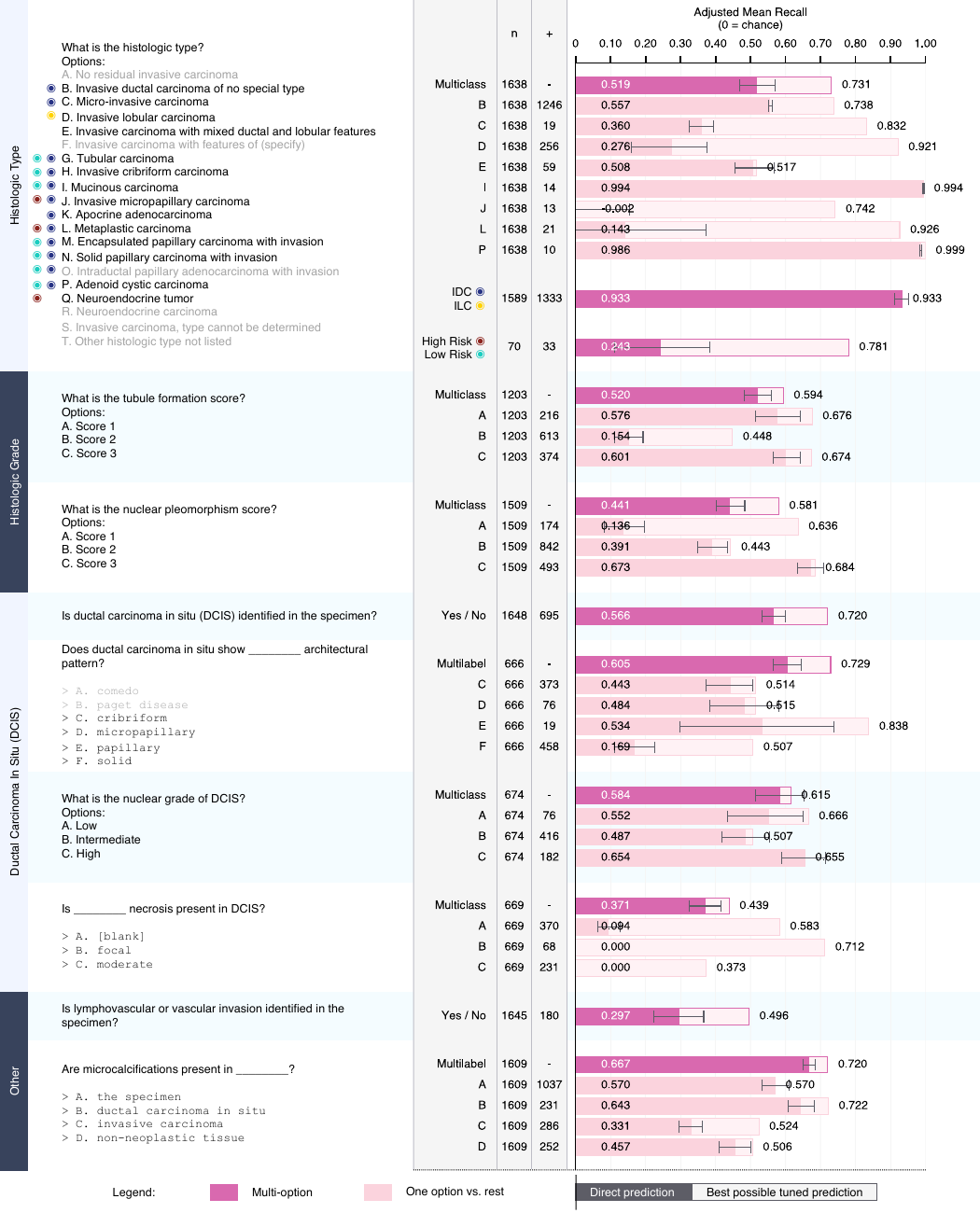}
        \phantomcaption
\end{figure*}
\begin{figure*}[!t]
    \ContinuedFloat
    \caption{
    Direct completion of pathology filling is enabled with dialogue. Multiple choice and yes-no \ac{QA} allows for directly filling a clinical report following \ac{CAP} guidelines, demonstrated here for invasive carcinoma of the breast (biopsy). Left: questions posed to PRISM2. For multiple choice questions (marked ``Multiclass''), all options are provided to the model at the same time. For yes-no questions (marked ``Yes / No'' (binary) or ``Multilabel''), each option in a multilabel report field completes the blank in the yes-no question. Right: predictive performance is measured with mean recall across all classes, adjusted such that randomly assigned values score 0 (eg. 0.5 is adjust to 0 for binary, 0.25 is adjusted to 0 for 4 classes). \Acf{AMR} is computed only across options with at least 10 examples in the evaluation data. \Ac{AMR} of the direct PRISM2 prediction is shown (number in bar) as a fraction of the maximum performance achievable by tuning per-class probability thresholds or weights (number right of bar). For histologic type, the gap between these is minimal when considering \ac{IDC} vs \ac{ILC} but large when considering high risk vs low risk types, suggesting that the relevant information is present in the model but its prediction is poorly calibrated for these classes. \textit{n} indicates number of samples and $+$ indicates number of positive samples. Error bars show the 95\% confidence interval computed with bootstrapping.
    }
    \label{fig:report_filling}
    \hrule
\end{figure*}

As a final demonstration of the value of direct prediction with dialogue, we aim to move beyond benchmark tasks, and evaluate PRISM2's ability to complete routine pathology reports, following \ac{CAP} reporting guidelines. We are the first to explore detailed pathology report completion directly with a pathology foundation model, without further training, and aim to provide a perspective on how a dialogue-based system could be used in real clinical practice.

We demonstrate this capability with the \ac{CAP} report template for biopsies with invasive carcinoma of the breast~\cite{CAPBreastInvasiveBx2023} (see Sec.~\ref{sec:eval} for details on the methodology and evaluation and Sec.~\ref{sec:eval_tasks} for dataset details). The questions associated with report fields and detailed performance metrics are shown in Fig.~\ref{fig:report_filling}. The fully shaded bars show the \acf{AMR} achieved directly by question-answering (``observed AMR''), whereas the lightly shaded extensions to these bars show the maximum \ac{AMR} achievable by calibrating the probabilities of the responses (``max AMR''). We can see that histologic type, \ac{DCIS} presence, \ac{DCIS} architectural pattern, and lymphovascular invasion fields benefit substantially from calibration. On the other hand, tubule formation score, \ac{DCIS} nuclear grade, necrosis, and microcalcifications benefit less. Overall, the results demonstrate that PRISM2 has modest to good understanding of the diagnostic concepts discussed but in some cases, the model's responses cannot be used without calibration.

Selecting the correct histologic type among 20 options is a challenging task. Only fourteen options were represented in the evaluation data, with eight options (B, C, D, E, I, J, L, P) having at least ten examples (see `n' and `+' in Fig~\ref{fig:report_filling}). PRISM2 achieved an \ac{AMR} of 0.519 (0.731 with calibration) across these options. Mucinous carcinoma (I) and adenoid cystic carcinoma (P) are the only options where the observed \ac{AMR} matches the max \ac{AMR} (0.994 and 0.986 observed \ac{AMR}, respectively); however, the observed \ac{AMR} for \ac{IDC} of no special type (B) and \ac{ILC} (D) is low (0.557 and 0.276, respectively). From the \ac{TCGA} BRCA results in Fig.~\ref{fig:direct_prediction_results}c and the breast \ac{IDC} and \ac{ILC} results in Extended Data Fig.~\ref{fig:extended_diagnostic_results}f, we know that these indications can be well identified by the model. It appears that correctly identifying these options becomes difficult when including all six ductal carcinoma variant options (B, C, I, J, L,  P), as well as the ductal and lobular mixed type (E). Indeed, a stratified analysis of histologic type classification reveals that \ac{IDC} vs. \ac{ILC} performance is good (0.933 \ac{AMR}) and does not benefit from calibration. On the other hand, when pooling rare histologic subtypes as prognostic of high risk or low risk, calibration is required. This serves as an example of a downstream task that could be developed from the concepts presented in this clinical report if validation (or training and validation) data are available.


\subsection{The diagnostic summary highlights key image regions}

Finally, connecting predictions to morphologic features can be an important tool for verifying model predictions. To this end, we selected two samples for detailed pathologic review. We qualitatively evaluate the generated text and compare with attention maps, the sum of all the scores per tile across attention heads and queries multiplied by the $\ell_2$-norms of the corresponding values.
As shown in Extended Data Fig.~\ref{fig:extended_roi_lymphoma} and~\ref{fig:extended_roi_gastric}, PRISM2 produces verbose descriptions of specimens. These descriptions remain consistent when looking only at the cropped regions that the model attends to the most. The model's attention maps, taken from the slide encoder's cross-attention layer, are shown for a single core of a \ac{WSI} and each of its top regions of interest.

\section{Discussion}

A complete pathology foundation model encodes the full range of relevant pathologic semantics not only within tiles, but also across \acp{WSI}. This minimizes the training and data required to adapt it to downstream tasks, while enabling generalization to diverse tasks like rare cancer detection and prognostic prediction. In this work, we achieve these aims. PRISM2 base embeddings outperform those of other slide-level models on diagnostic, prognostic, and biomarker tasks. PRISM2 diagnostic embeddings, which benefit from a language model trained on diagnostic dialogue, further improve performance on diagnostic tasks. Indeed, PRISM2 is the only model to match or exceed the performance of all tested clinical-grade products in cancer detection and does so with direct question answering without further training or adaptation. Finally, PRISM2 survival embeddings, trained on an overall survival cohort of unprecedented scale, containing nearly 100 thousand patients, further outperform the base embeddings on prognostic tasks including \ac{RFS} and \ac{DSS}.

Building on a language model was critical for unlocking PRISM2 direct prediction performance. Natural language provides a straightforward manner in which to frame clinical tasks, even when the positive or negative cases are hard to define. For example, contrastive direct prediction for pan-cancer detection requires all possible cancer-positive phrases to be contrasted with all possible non-cancer phrases, including benign conditions across all tissues (see Supplementary Tab.~\ref{tab:supp_contrastive_pancancer_prompts}). Conversely, the same yes-no question (e.g. ``Is malignant tumor present in this specimen?'') can be used across all pan-cancer data. This consideration may be less important when the number of candidate labels is small, as is often the case for subtyping. However, we note that we do not observe contrastive approaches outperforming PRISM2 direct prediction in these cases, thereby indicating that PRISM2 allows for general \ac{QA} without sacrificing performance on more narrowly defined tasks. Moving to linear classification removes this consideration, but PRISM2 still outperforms competing models.

Optimization details were also important for achieving generalizability. We freeze the slide encoder early in training before reaching the lowest contrastive validation loss. We find this achieves a better balance between performance on diagnostic tasks and transferability to biomarker and prognostic tasks, in line with other work that also has observed that earlier checkpoints may be more generalizable~\cite{Evci2022Head2Toe}. We then rely on the vision-language model to refine the representation into one specialized for diagnosis. We have observed freezing the slide encoder prevents the vision-language model from overfitting and ultimately improves performance.

Although PRISM2 is capable of dialogue, accurate question answering is prioritized over robust chat capabilities. While abilities beyond our training set, like answering compound questions, are inherited from the Phi-3-Mini initialization, they are not used for downstream tasks. Instead, training the model to answer yes-no and multiple-choice questions, alongside open-ended questions and more general report generation provides us with three ways in which the vision-language model of PRISM2 can be adapted to downstream tasks: (1) by uncalibrated direct prediction, (2) by direct prediction of probabilities calibrated on a validation set, and (3) by further training using base or diagnostic embeddings.
The first setting allows direct prediction of conditions without further training. Additionally, instead of directly using the PRISM2 responses, we can quantify the probability of response to a specific yes-no and multiple-choice question. Finally, the diagnostic embeddings are tuned for diagnostic pathology tasks such as the detection and identification of cancers, precursors to cancer, and benign conditions and can be adapated when additional training and validation data are available.

There are still open questions regarding directions of possible improvement for slide-level foundation models. Because the perceiver-based slide encoder relies on cross-attention rather than self-attention, we did not include position encoding. Also, although Virchow2 can encode tiles at many magnifications, PRISM2 only used embeddings at 0.5 \ac{mpp}. Moving to a different aggregator architecture with position encoding and mixed magnification input may improve performance on some tasks that benefit from greater spatial context, like cancer subtyping, and may enable measurement tasks, like mitotic event counting. As evidenced by the modest improvement in biomarker performance compared to baseline models, as well as the large improvement in survival prediction performance gained by fine-tuning on \ac{OS} data, we posit that further improvement in prognostic and biomarker tasks will require including them in the model pre-training phase. Finally, while it appears that full diagnostic report completion is possible, it remains challenging. A complete study of report completion would evaluate inter-pathologist variability. Grading and subtyping are particularly subjective and may be subject to biases.

 Work in foundation models in computational pathology has moved from the tile level to the slide level~\cite{shaikovski2024prism,ding2024multimodal,wang2024pathology,xu2024multimodal,jaume2024transcriptomics,jaume2024multistain,vaidya2025molecular,ahmed2024pathalign}. PRISM2 extends the scale of this effort and is the first to achieve clinical-grade cancer detection performance. It is the largest model (4.6 billion parameters) trained on the most comprehensive dataset to date with nearly 700 thousand specimens and clinical reports, comprising over 2.3 million slides and 14M question-answer pairs. Besides the scale, we demonstrate through the inclusion of a large vision-language model that the image-report alignment methods to date can be viewed as a base for further foundation model development, and model performance can be further improved by borrowing from the \acf{LLM} domain. Finally, we find that diagnostic performance in pathology is beginning to saturate and that further improvements in biomarker and prognostic prediction, driven by the inclusion of this data in foundation model pre-training, may be the new frontier for foundation models in pathology.

\ifnum\papermode=0
\section{Online Methods}
\else
\section{Methods}
\fi
\label{sec:methods}

\subsection{Training Data} \label{sec:method_data}
Institutional review board review was not applicable for the research
described in this study. This research study was conducted retrospectively from deidentified data licensed to Paige.AI, Inc. from \ac{MSK}. The data used in this study were all collected originally for clinical use by \ac{MSK} in the practice setting and are therefore considered secondary data. Only data previously deidentified by \ac{MSK} were utilized in the analysis, and unique patient identifiers were completely removed from the analytical dataset. To the best of our knowledge, \ac{MSK} has not transferred any data for which the applicable patient has not consented to or otherwise agreed to \ac{MSK}’s Notice of Privacy Practices or a substantially similar notice, waiver or consent.

The training data is comprised of 685,507 specimens collected from 200,692 patients. Each specimen is associated with a report and one or more whole slide images, totaling 685,507 reports and 2,3510,518 slides. The \acp{WSI} were stained with routine hematoxylin and eosin stain and scanned at 20$\times$, 0.5 microns per pixel, using Leica scanners. The data represents a diverse collection of tissues, diagnoses, and medical centers, all reported at \ac{MSK},(see Fig.~\ref{fig:data}). Each modality is preprocessed to prepare for model training.

\acs{WSI}s are processed using a fully-convolutional network to perform tissue segmentation and tiled at $20\times$ magnification with a size of $224\times224$ pixels, discarding any tiles that are less than 65\% tissue. The resulting tiles are then processed by Virchow2~\cite{zimmermann2024virchow2} to produce tile embeddings, specifically the 1280-dimensional class token output of the model. During training, up to 100K tiles are loaded per specimen. In cases where specimens have more than 100K tiles, \acp{WSI} in the specimen are dropped until the criterion is met. This restriction does not apply during inference.

Raw report data is delivered at the case-level, but reporting is often done at the specimen-level which requires extracting and matching each report to the proper set of \acs{WSI}s for that specimen. The reports are a mix of synoptic worksheets and free-text and are stripped of boilerplate content like headers/footers, disclaimers, and repetitious formatting characters.

\subsection{Creating clinical dialogue for training} \label{sec:method_rewrites}
To create high-quality text data for training, we apply GPT-4o~\cite{openai2024gpt4ocard} with various prompting strategies to the report data (see Fig.~\ref{fig:data} for an example workflow). We create four classes of processed text tasks to be used in training:
\begin{enumerate}
    \item \textbf{Clinical report generation.} Given a task string (i.e. ``Write a report''), the model must generate the clinical report.
    \item \textbf{Yes-No question answering.} Given a binary user question, the model must generate an answer.
    \item \textbf{Open-ended question answering.} Given an open-ended user question, the model must provide an answer.
    \item \textbf{Multiple-choice question answering.} Given a user question and a list of answer options, the model must select the correct option.
    \item \textbf{Image-text matching.} Given a report, the model must identify if the report accurately describes the given \acp{WSI}, and then generate the correct report if they are mismatched.
\end{enumerate}
We choose these templates to broaden the training distribution beyond the obvious report generation task and enable dialogue-based interactions with the trained model. The process to create the data for each of these tasks is described below and the full system prompts can be found in Supplementary Note~\ref{sec:apx_gpt4_rewrites}.

\textbf{Clinical report rewrites.} Using the synoptic reports and free-text notes, we prompt GPT-4o to rewrite all input text, including patient history and specimen description when available, into natural language sentences while retaining all diagnostically relevant findings. 

\textbf{Yes-No question-answer pairs.} Question-answer pairs are generated through a multi-step process. First, GPT-4o is prompted to convert clinical reports into lists of individual findings, similar to MAIRA-2~\cite{bannur2024maira2}, which are then filtered to exclude molecular and \ac{IHC} results. These filtered findings are then used as input to GPT-4o with a prompt to generate question-answer pairs for each finding that can be answered with simply ``Yes'' or ``No''. While effective, we found this process is naturally biased towards creating questions that have the answer ``Yes.''. We expect that this is because negative findings are typically not mentioned in the reports. For instance, a report positively identifying \ac{DCIS} in breast tissue allows GPT-4o to generate a pair such as: ``Q: Is DCIS present? A: Yes.'' However, a normal breast tissue report does not mention \ac{DCIS}, so no negative question-answer pair of the form ``Q: Is DCIS present? A: No'', is generated, biasing the dataset (and resulting model) toward affirmative responses. Inspired by Goyal et al.~\cite{goyal2017vqa}, to mitigate this bias, we implemented a simple method to mine complementary images for additional questions. Specifically, we pair random reports with random questions and prompt GPT-4o to answer based on the provided report. This yields complementary image-question-answer triplets, balancing the data distribution. 

\textbf{Open-ended question-answer pairs.} Open-ended question-answer pairs are also generated from the list of findings described above. We instruct GPT-4o to create the \ac{QA} pairs such that the questions are not answerable with just yes/no but require more specific answers. Open-ended \ac{QA} pairs help mimic more realistic model interactions and improve the diversity of the training data.

\textbf{Multiple-choice question-option-answer triplets.} Multiple-choice question-options-answer triplets are generated from the synoptic reports and free-text notes. We prompt GPT-4.1~\cite{openai2025gpt41} to first classify a given report into one of 90 possible \ac{CAP} Cancer Protocol Templates and then to create as many question-options-answer triplets as possible from the report such that they follow the inferred protocol. We only provide the name of the 90 possible protocols and not the full template as we rely on the inherent recall of these protocols by GPT-4.1.

\textbf{Image-text matching.} The matching task uses the clinical report rewrites described above and does not require further preprocessing. This task aims to force the data to use the image information to complete the task.

\subsection{PRISM2 architecture}
\label{sec:arch}
The overall architecture of PRISM2 consists of a slide encoder, a language encoder, and a language model, as illustrated in Extended Data Fig.~\ref{fig:arch} and described below.

\textbf{Slide encoder.} The slide encoder uses a perceiver~\cite{jaegle2021perceiver} architecture. The PRISM2 perceiver uses 1 block, defined as a cross attention layer between latent queries and tile embeddings followed by 6 self-attention transformer layers for the latent queries. ac{GQA}~\cite{ainslie2023gqa} is used in the tile embedding cross attention. The number of latent queries is set to 256 and an attention pooling layer is used on these to produce the embedding used in the contrastive objective. This embedding is referred to as the \textit{base} embedding. The perceiver and attention pooler have 541M and 79M parameters, respectively.

\textbf{Language encoder.} BioGPT~\cite{luo2022biogpt} is used as the language encoder. 
We initialize a learnable class token which is concatenated with the input text sequence. The final layer hidden state of the class token is used as the text embedding for the contrastive objective after a linear projection and $\ell_2$ normalization.

\textbf{Large language model decoder.} For the dialogue-focused language model component we use Phi-3 Mini~\cite{abdin2024phi}, a 3.8B parameter decoder-only \ac{LLM}. We adopt a LLaVa-style~\cite{liu2023visual} approach to enabling Phi-3 Mini to process both images and text. The 256 latents outputted by the slide encoder are further processed by a 2-layer \ac{MLP} adapter (29M parameters) to project them into the Phi-3 embedding space and are inserted into the text token embedding sequence as input to the \ac{LLM}. This allows Phi-3 to integrate the slide and text information via its self-attention layers. In addition to decoding the language tokens, we also perform analysis with the Phi-3 Mini final layer hidden state of the \texttt{<|assistant|>} token. We refer to as the \textit{diagnostic} embedding.

\subsection{PRISM2 training}
\label{sec:method_training}
\textbf{Objectives.} PRISM2 is trained with both contrastive and autoregressive objectives. Following~\cite{yu2022coca,shaikovski2024prism}, the contrastive loss is 
\ifnum\papermode=0
\begin{multline}
    \mathcal{L}_{\text{con}} = -\frac{1}{N} \left ( \sum_{i=1}^{N} \log \frac{\exp(\boldsymbol{v}_i^\top \boldsymbol{t}_i/\tau)}{\sum_{j=1}^{N} \exp(\boldsymbol{v}_i^\top \boldsymbol{t}_j/\tau)} \right. \\
    +\left. \sum_{i=1}^{N} \log \frac{\exp(\boldsymbol{t}_i^\top \boldsymbol{v}_i/\tau)}{\sum_{j=1}^{N} \exp(\boldsymbol{t}_i^\top \boldsymbol{v}_j/\tau)}\right ),
\end{multline}
\else
\begin{equation}
    \mathcal{L}_{\text{con}} = -\frac{1}{N} \left ( \sum_{i=1}^{N} \log \frac{\exp(\boldsymbol{v}_i^\top \boldsymbol{t}_i/\tau)}{\sum_{j=1}^{N} \exp(\boldsymbol{v}_i^\top \boldsymbol{t}_j/\tau)} \right. 
    +\left. \sum_{i=1}^{N} \log \frac{\exp(\boldsymbol{t}_i^\top \boldsymbol{v}_i/\tau)}{\sum_{j=1}^{N} \exp(\boldsymbol{t}_i^\top \boldsymbol{v}_j/\tau)}\right ),
\end{equation}
\fi
where $N$ is batch size, $\tau >0 $ is a learned temperature, and $(\boldsymbol{v}_i, \boldsymbol{t}_i)$ are the image and text representations, respectively, as defined above. Only the diagnostic report rewrites are used as the text input for the contrastive objective. For the autoregressive objective, we apply the standard cross-entropy loss on next-token predictions from Phi-3 Mini. We use the chat templates as the text input and only apply the loss to the \texttt{assistant} regions of the templates:

\begin{equation}
    \mathcal{L}_{\text{chat}} = -\sum_{t \in \mathcal{A}} \log p(y_t \mid y_{<t}, \boldsymbol{X}),
\end{equation}
where $\mathcal{A}$ is the set of token indices corresponding to assistant tokens and $\boldsymbol{X}$ are the image latents after the \ac{LLM} adapter \ac{MLP}. The losses are linearly combined to create the total loss:

\begin{equation}
    \mathcal{L}_{\text{total}} = \lambda_{\text{con}}\mathcal{L}_{\text{con}} + \lambda_{\text{chat}}\mathcal{L}_{\text{chat}}.
\end{equation}

\textbf{Data sampling.} For the contrastive objective, report rewrites are randomly sampled from five rewrites per report. For the autoregressive objective, we adopt the Phi-3~\cite{abdin2024phi} chat template using alternating \texttt{user} and \texttt{assistant} tags. The first \texttt{user} message always contains the \ac{WSI} latents from the slide encoder followed by an initial prompt. In each template, the patient history and specimen description are randomly included, each with a probability of $20\%$, to allow for conditioning the model on this extra context without being overly reliant on it. Complementary \ac{QA} pairs are included with a sampling rate of $20\%$. In total, we create 685K report rewrites for as many specimens and 14M question-answer pairs.

\textbf{Optimization.} We adopt a two stage training approach. In the first stage the slide encoder, attention pooler, \ac{LLM} adapter, and BioGPT weights are updated using the full objective. The Phi-3 Mini weights are constant (frozen) during this stage. 
We set $\lambda_{\text{con}}$ to $0.25$ and $\lambda_{\text{chat}}$ to $1.0$. In the second stage, the contrastive objective is discarded and the slide encoder weights are frozen. The \ac{LLM} adapter and Phi-3 Mini are updated to allow the language model to better integrate the perceiver latents and model the language of pathology. The optimizer and learning rate schedules are reset for this stage.

We train the model on 56 A100 40GB GPUs with bf16 precision. We use the AdamW~\cite{loshchilov2019dadamw} optimizer with $\beta_1=0.9$ and $\beta_2=0.95$ and a learning rate (LR) warmup for 500 steps which is then held constant. We do one epoch of stage 1 training with a post-warmup LR of $1.0 \times 10^{-4}$ on all trainable components except the language encoder which uses an LR of $2.0 \times 10^{-5}$, and a 0.1 decay on the slide encoder weights. The large language encoder (Phi-3-Mini) weights are kept frozen in stage 1 but all dialogue objectives are applied. We do 16 epochs of stage 2 training with 500 steps of LR warmup to $2.0 \times 10^{-5}$ and no weight decay. The slide and language encoders (perceiver and BioGPT) are frozen and the large language encoder (Phi-3-Mini) is unfrozen for fine-tuning. An epoch is defined as one pass over each of the four chat templates per specimen, for all specimens in the training data. Because the contrastive objective aligns the image representation with the report and not with each chat template, each specimen is paired with its report four times in a single epoch. Several engineering changes were required for efficient training (see Supplementary Note~\ref{sec:apx_eff_imp} for details). 

\textbf{Survival fine-tuning.} The PRISM2 slide encoder was fine-tuned on a large dataset of 225,597 cases with \ac{OS} event data, containing 69,114 death events. Survival prediction is trained by learning a new cross-attention pooling in the (perceiver) slide encoder for ``survival'' embeddings instead of ``base'' embeddings and adding a linear regression head on top. This head predicts the log relative hazard, along with the event time in months, for the partial log likelihood loss, completing the Cox proportional hazards model. We used the AdamW~\cite{loshchilov2019dadamw} optimizer with $\beta_1=0.9$ and $\beta_2=0.98$ and a learning rate (LR) warmup for 4000 steps which is then held constant. We used a post-warmup  LR of $1.0 \times 10^{-4}$ for the slide encoder and $1.0 \times 10^{-3}$ for the prediction head, with a 0.0001 weight decay. The comparison survival specialist model was trained from scratch using the same architecture with the same hyperparameters as stage 1 PRISM2 training.

\subsection{Evaluation methods} \label{sec:eval}

\textbf{Direct prediction with question answering}. Direct prediction with question answering refers to a setting without additional training and includes multiple-choice and yes-no settings. 
With both yes-no and multiple-choice \ac{QA}, we can quantify the probability of predictions. For each direct classification task, we choose a question prompt $w_{<i}$ (e.g. \textit{Is malignant tumor present?}) and consider a set $C$ of possible completions $c$. For yes-no \ac{QA}, these are `Yes' or `No'; for multiple-choice \ac{QA}, these are the letter options (eg. `A', `B').

Formally, we define our dialogue-based direct predictions as
\ifnum\papermode=0
\begin{multline}
    f(w_{<i}, x) = \argmax_{c \in C} \log P(w_i = c | w_{<i}, x) \\ - \log P(w_i = c | w_{<i}),
\end{multline}
\else
\begin{equation}
    f(w_{<i}, x) = \argmax_{c \in C} \log P(w_i = c | w_{<i}, x) - \log P(w_i = c | w_{<i}),
\end{equation}
\fi
where $c$ is a completion in a set of completions $C$ that cover all classes in the task, $w_{<i}$ is a tokenized prompt containing a question and $x$ is the context which contains \ac{WSI} embedding tokens from the slide encoder.

\textbf{Contrastive direct prediction by ranking similarity scores}. We evaluate direct prediction with PRISM and TITAN by reformulating their contrastive image-text alignment objectives as multi-class classification. Given a specimen-level embedding $\boldsymbol{v}$ from the slide encoder and a set of prompt embeddings $\boldsymbol{t} \in \boldsymbol{T}$ from the language encoder, one per each class, we compute the cosine similarity between each prompt embedding and apply a softmax activation to the resulting logits.
Each entry in the vector corresponds to a probability of that prompt matching the slide.
In the case of multiple per class, we take the prompt with the highest similarity score as the prompt representing that class.
Formally, the probability of the class $i$ being the true class is

\begin{equation}
    P(C=i \mid \boldsymbol{v}) = \frac{\exp(\boldsymbol{v}^\top \boldsymbol{t}_i)}{\sum_{\boldsymbol{t}_j \in \boldsymbol{T}} \exp(\boldsymbol{v}^\top \boldsymbol{t}_j)}.
\end{equation}

\textbf{Linear probe}. Beyond the direct setting, model embeddings are also evaluated by training a linear classifier or linear regression. To do so, model embeddings are generated for training, validation, and testing sets of data. Linear classifiers are trained to convergence for each of 24 L2 regularization weights, with the best classifier selected on the validation set. For survival analysis, a Cox proportional hazards model is fitted to the training set for each of 24 elastic regularization weights (L1 and L2, with an L1:L2 ratio of 0.0001), with the best regressor selected on the validation set.

\textbf{Report completion}. We demonstrate the completion of pathology reports with the \ac{CAP} worksheet for biopsies with invasive carcinoma of the breast~\cite{CAPBreastInvasiveBx2023}. The specimen procedure and laterality fields and the tumor position field are omitted from this analysis, as they constitute sample collection information rather than diagnostic findings derived from \ac{HE} stained slides. The mitotic rate measurement is also excluded as the model was not trained to count. Filling the fields in the \ac{CAP} template is posed as multiple-choice \ac{QA} for multiclass fields (eg. histologic type, histologic grade) or as yes-no \ac{QA} for binary and multilabel fields (eg. presence of \ac{DCIS} or its architectural pattern). The questions and detailed performance metrics are shown in Fig.~\ref{fig:report_filling}. In total, the report can be represented with nine sets of fields, grouped under histologic type, histologic grade, \ac{DCIS}, and ``other'' (lymphovascular invasion and microcalcifications), filled by asking 5 multiple-choice questions and 9 yes-no questions. We measure the \acf{AMR}. For multiclass, this is equivalent to adjusted balanced accuracy (adjusted such that random chance is scored as 0).

For direct prediction with PRISM2, we infer a prediction probability as the probability of the selected option or of `yes' vs. `no'. This allows us to evaluate whether the model predictions can be used directly or if they require further calibration, as is often done with clinical-grade products. We evaluate both the \ac{AMR} of direct prediction (observed) and the best possible \ac{AMR} achieved by calibrating the probability (max). Calibration is per-class thresholding for binary (and multilabel) tasks, or per-class reweighting for multiclass tasks. We find the best thresholds by sweeping thresholds per class and the best reweighting with numerical optimization.

\subsection{Comparison models} \label{sec:comp_models}
Several slide- or part-level embedding models have previously been reported and are used for comparison. We describe each briefly here. PRISM~\cite{shaikovski2024prism} is the previous model developed by the PRISM2 team. It differs from PRISM2 in that it is trained using few samples, with fewer types of text outputs and a different training algorithm. TITAN~\cite{ding2024multimodal} is a slide-level embedding model that uses vision-language pretraining. TITAN aggregates CONCH~\cite{lu2024visual} in a three-stage processing using both synthetic captions and medical reports and a training dataset of approximately 340 thousand whole slide images. COBRA~\cite{lenz2025unsupervised} uses tile representations from four different foundation models to serve as feature space augmentations for a contrastive self-supervised learning approach. COBRA aggregates tiles using a state-space architecture and is trained using approximately 3 thousand slides. Prov-Gigapath~\cite{xu2024whole} uses a vision-only approach to creating a slide-level model. Tile embeddings are aggregated using a masked auto-encoding approach and a training dataset of approximately 170 thousand whole slide images. Virchow2 Mean refers to a baseline where the Virchow2~\cite{zimmermann2024virchow2} tile embeddings are naively aggregated by simply averaging them all together. Both PRISM2 and COBRA aggregate Virchow2 tile embeddings.

\subsection{Evaluation datasets} \label{sec:eval_tasks}

We evaluate the embeddings of PRISM2 and baseline models on a range of cancer detection, cancer subtyping, and molecular tasks described below.

\textbf{Pan-cancer Detection} is an in-house evaluation dataset containing 22,932 \ac{HE} \acp{WSI} collected from 6,142 specimens. Samples are scanned at \acf{MSK} and sourced in near-equal quantities from \ac{MSK} (49\%) and \acp{WSI} submitted to \ac{MSK} for review from globally diverse external sites (51\%). Of the 16 cancers included in the dataset, seven are rare. The National Cancer Institute (NCI) defines rare cancers as those with an annual incidence in the United States below 15 people per 100,000. Complimenting the evaluation dataset, the training and validation sets together contain 89,417 slides across 40,402 specimens and follow a more natural distribution for the frequency of different cancers. Note that the linear classifier trained on the pan-cancer training set is used to evaluate invasive cancer detection via linear probing on the prostate, breast, and breast lymph node product benchmarks below.

\textbf{Prostate product benchmark.} This dataset contains 2947 blocks (3327 slides) of prostate needle core biopsies. Labels for the blocks are extracted from synoptic reports collected at \ac{MSK}. This dataset has been curated to evaluate the standalone performance of Paige Prostate Detect, which is a tissue-specific, clinical-grade model. We use this dataset to evaluate invasive cancer detection. 

\textbf{Breast product benchmark.} This dataset contains 190 slides with invasive cancer and 1501 benign slides, labelled individually by a pathologist according to presence of \ac{ADH}, \ac{ALH}, \ac{LCIS}, \ac{DCIS}, \ac{IDC}, \ac{ILC} and/or other subtypes. This dataset has been curated to evaluate the standalone performance of Paige Breast, which is a tissue-specific, clinical-grade model. We group \ac{IDC} and \ac{ILC} into an invasive cancer label to evaluate invasive cancer detection.

\textbf{Breast lymph node product benchmark.} This dataset contains 458 lymph node slides with metastasized breast cancer, and 295 benign lymph node slides. Each slide has been labeled by a pathologist according to presence of invasive cancer, and largest tumor on the slide is measured to categorize the tumor into macrometastasis, micrometastasis, or \acp{ITC}. We group the categories together into an invasive cancer label to evaluate invasive cancer detection.

\textbf{Breast Subtyping} is an in-house dataset comprising 11,774 breast \ac{HE} \acp{WSI} spanning 2407 specimens. Each specimen is associated with six non-mutually exclusive labels indicating the presence or absence of invasive lobular carcinoma (ILC), invasive ductal carcinoma (IDC), ductal carcinoma in situ (DCIS), lobular carcinoma in situ (LCIS), atypical ductal hyperpalasia (ADH), and atypical lobular hyperplasia (ALH). The dataset is divided at a specimen level into 8 folds with a 75/12.5/12.5 percentage split between train, tune, and test subsets.

\textbf{MSK GI} is an in-house dataset containing 226,024 gastrointestinal (GI) \ac{HE} \acp{WSI} gathered from 122,601 specimens. The labels span 12 non-mutually exclusive binary classification tasks. The dataset is split into train, tune, and test subsets of size 91276, 17515, and 13810 specimens respectively. 

\textbf{TCGA-BRCA} is a publicly available invasive breast cancer (BRCA) dataset focused on discriminating between IDC and ILC. It comprises 1002 \ac{HE} \acp{WSI} from TCGA. The data is split into five folds with a 60/20/20 ratio between the train, tune, and test subsets for each fold. 

\textbf{TCGA-NSCLC} is a publicly available non-small cell lung cancer (NSCLC) dataset composed of 1043 \ac{HE} \acp{WSI} from TCGA. Each \ac{WSI} contains lung adenocarcinoma (LUAD) or lung squamous cell carcinoma (LUSC). The data is split into five folds with a 60/20/20 ratio between the train, tune, and test subsets for each fold. 

\textbf{TCGA-RCC} is a publicly available renal cell carcinoma (RCC) dataset containing 938 \ac{HE} \acp{WSI}. Each slide belongs to one of three ground truth subtypes: clear cell renal cell carcinoma (ccrcc), papillary renal cell carcinoma (prcc), or chromophobe renal cell carcinoma (chrcc).  The data is split into five folds with a 60/20/20 ratio between the train, tune, and test subsets for each fold. 

\textbf{MSK biomarker prediction} is an in-house dataset containing 37,633 \acp{WSI} gathered from 33,949 samples. The ground truth labels span 10 different biomarkers: Prostate AR, Ovarian FGA, Esophagogastric HER2, Colorectal MSI, Lung EGFR, Melanoma BRAF, Bladder FGFR, Breast CDH1, Endometrial PTEN, and Breast PIK3CA. These labels were originally identified by MSK-IMPACT \cite{cheng2015-msk-impact}, a targeted test for genetic mutations. The dataset is split into train, tune, and test subsets of size 22,634, 5,606, and 5,709 specimens respectively.

\textbf{TCGA biomarker prediction} is a public dataset containing 10,493 \acp{WSI} gathered from 8,611 samples. The ground truth labels are a subset of those in the \ac{MSK} biomarker prediction dataset: Colorectal MSI, Lung EGFR, Melanoma BRAF, Bladder FGFR, Breast CDH1, Endometrial PTEN, and Breast PIK3CA. Training for each label is done with 3-fold cross-validation.

\textbf{MSK CRC RFS} is an in-house dataset of \ac{CRC} cases curated for recurrence analysis and \ac{RFS} prediction. It contains 884 recurrence events across 1,260 cases, spanning 6,513 \acp{WSI}. Evaluation on this data is done with 5-fold cross-validation, using month-level precision for the time to event.

\textbf{TCGA CRC DSS} is a set of 13 tasks derived from the public \ac{TCGA} dataset spanning 9,041 cases with 11,072 \acp{WSI} and 2,577 disease-specific death events. \ac{DSS} is evaluated with 5-fold cross-validation for each of the following tasks: CRC (colorectal cancer: COAD and READ), NSCLC (non-small cell lung carcinoma: LUAD and LUSC), RCC (renal cell carcinoma: KIRK, KIRP, and KICH), BRCA (breast cancer), CNS (central nervous system: GBM and LGG), SARC (sarcoma), ENDO (endometrial: UCEC and UCS), LBTC (liver and biliary tract cancer: LIHC and CHOL), MEL (melanoma: SKCM and UVM), HNSC (head and neck cancer), BLCA (bladder cancer), GI (upper gastrointestinal tract: ESCA and STAD), and PAAD (pancreatic cancer). Tasks were curated such that they have at least 100 \ac{DSS} events each.

\textbf{Invasive Breast Cancer Pathology Report Completion} is an in-house dataset of 1,638 invasive breast cancer biopsy specimens and their associated pathology reports. The pathology reports are standardized to match the current \ac{CAP} ``Breast Biopsy, Invasive'' cancer protocol template. Specimens were selected to ensure the histologic type of the invasive cancer is represented by one of the closed-ended options in the protocol. Macroscopic and quantitative protocol fields were omitted from evaluation. Specifically, procedure type, specimen laterality, tumor site, mitotic rate, and tumor measurement were not included.

\textbf{Metrics.} For linear probing, we use (one-vs-one) \acf{AUC} as the primary metric so that model performance could be evaluated without tuning a threshold for the predicted class probabilities. On the other hand, we primarily evaluate direct predictions with balanced accuracy by assuming a threshold at 0.5 probability. For question answering, this is equivalent to taking the model responses directly, without calibration. For fairness, we use product-specific thresholds for clinical-grade products, as these were carefully tuned during product development. Balanced accuracy is the mean recall across all classes which makes it a suitable metric for imbalanced data.

\textbf{Statistical analysis.} Statistical significance was assessed using two-sided paired permutation tests with 1,000 permutations. Confidence intervals of 95\% were estimated via bootstrapping using 1,000 iterations. In cases where direct prediction is also reported, metric computation (and thus statistical analyses) are done with the micro-average across folds to enable direct comparison (i.e. no class-reweighting). For TCGA DSS, TCGA biomarkers, and MSK biomarkers the macro-average is used.


\ifnum\papermode=0
\subsection{Software}
All data processing, model training, and analysis were performed using Python (v.3.10) and open-source libraries. For data curation, we used Pandas (v.2.2.2) for metadata management, and OpenSlide (v.1.3.1) and Pillow (v.10.0.0) for WSI preprocessing and image decoding. Specimen- and report-level metadata were managed through internal Paige libraries. All model development and experiments were conducted using PyTorch (v.2.3.1) and PyTorch Lightning (v.2.3.0) as the core deep learning frameworks.
We used the publicly available \href{https://huggingface.co/paige-ai/Virchow2}{Virchow2} model for WSI embedding generation.  used and torchsurv (v.0.1.4) for survival-specific objectives. Additional scientific libraries included scikit-learn (v.1.4.2), einops, and nltk (v.3.9). Dialogue generation and evaluation were implemented through the Paige internal reporting framework using the OpenAI API (openai v.0.28.1) with GPT4o model.

\subsection{Reporting summary}
Further information on research design is available in the Nature
Portfolio Reporting Summary linked to this article.

\subsection{Data availability}
This study did not specifically collect patient data. The retrospective
analysis utilized proprietary deidentified digital pathology whole
slides and associated metadata were exclusively licensed by Paige.AI,
Inc. from MSKCC. Requests for data need to be submitted to Paige AI
(https://paige.ai/contact-us/) and evaluated by Paige AI and MSKCC on a
case-by-case basis. All requests complying with internal regulations on
data privacy and intellectual property will be granted. All TCGA data used in this study are publicly available through the NCI Genomic Data Commons (GDC) portal at \url{https://portal.gdc.cancer.gov/}.

\subsection{Code availability}
The PRISM2 training and inference pipelines were developed using internal Paige AI and Microsoft Research infrastructure and cannot be fully released due to dependencies on proprietary libraries, distributed compute environments, and commercial intellectual property complexities.
A zip archive containing all performance-analysis and figure-generation scripts used to produce the results in this study is included with this submission and will be made publicly available on GitHub upon publication.
Academic users will be able to access PRISM2 embeddings through the Paige Inference API for non-commercial research use on their own cohorts, subject to a data-use agreement and approval by Paige AI.
All experiments are documented in the Methods section to enable independent replication.
PRISM2 builds upon open-source components including the \href{https://github.com/pytorch/pytorch}{PyTorch} and \href{https://github.com/Lightning-AI/pytorch-lightning}{PyTorch Lightning} frameworks.
\fi

\ifnum\papermode=0
\FloatBarrier
\bibliography{references}
\onecolumn
\fi

\section{Acknowledgments}
We gratefully thank Philip Rosenfield and Djamilia Dierov for their contributions in making this collaboration possible, Philippe Mathieu for distributed inference support, Mark Fleishman and Sid Senthilnathan for data support, and Wayne Hendricks and Alexander van Eck for infrastructure support.

\ifnum\papermode=0
\else
\FloatBarrier
\printbibliography
\fi

\ifnum\papermode=0
\section{Author contributions}
E.V., G.S., A.C., J.V., E.Z. and K.S. wrote code, developed infrastructure, trained models, and analyzed results throughout the study. S.L., K.S., R.Y., N.F., and T.J.F. contributed to study conception and coordination, building a cross-company partnership and research team, establishing data-sharing agreements, and performed evaluation and analysis. N.T. provided technical guidance and reviewed algorithmic aspects of the study. J.A.R., J.S., M.G., D.S.K and M.R.W. provided clinical guidance and reviewed the clinical aspects of the study; J.S., M.G., M.R.W., D.S.K specifically formulated the colorectal cancer recurrence problem and curated the corresponding cohort. E.V., G.S., A.C., Y.K.W., R.A.G., and J.H.B. worked on data preparation, with Y.K.W., J.H.B., and R.A.G. focusing on biomarker data preparation and benchmark construction. E.V., G.S., A.C., J.V., E.Z. K.S. and S.L. wrote the initial draft and all authors provided feedback on the manuscript.

\section{Competing interests}
E.V., G.S., A.C., J.V., R.Y., Y.K.W., R.A.G., J.H.B., T.J.F., and S.L. are current or former employees of Paige.AI, Inc. and hold equity in Tempus AI. E.Z., N.T., N.F., and K.S. are employees of Microsoft. S.L., E.V., G.S., A.C., J.H.B., R.A.G., and T.J.F. are inventors on a provisional U.S. patent application (No. 18/521903) related to methodological aspects of this work. The remaining authors declare no competing interests.
\fi

\renewcommand{\figurename}{Extended Data Figure}
\setcounter{figure}{0}

\clearpage
\begin{figure*}[t]
    \centering
    \begin{subfigure}[b]{\textwidth}
         \centering
         \includegraphics[width=\textwidth]{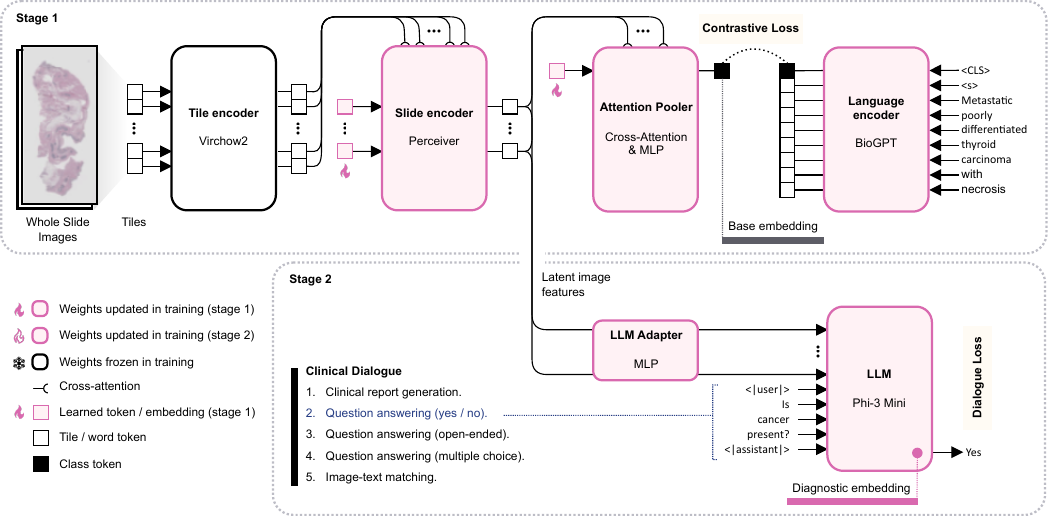}
     \end{subfigure}
    \caption{
    The architecture and training schematic for PRISM2. Sets of whole slide images are summarized as a base embedding (Attention Pooler class token) which is aligned to the representation of the clinical report diagnostic summary (encoded by the language decoder, BioGPT) by minimizing a contrastive loss. The image latent features are also passed to a pretrained large language model (\ac{LLM}, Phi-3 Mini) along with a prompt from one of four dialogue templates, with the model's response updated by minimizing the dialogue loss. Training proceeds in two stages: in stage 1, only the slide encoder, attention pooler, language decoder, and \ac{LLM} adapter weights are updated; in stage 2, only the \ac{LLM} adapter and \ac{LLM} weights are updated. For use in downstream tasks, the latent image features produced by the slide encoder are summarized in two embeddings: (1) the features are pooled into a `base embedding', used for contrastive alignment; and (2) the features are refined by the \ac{LLM} into a `diagnostic embedding', taken from the \ac{LLM} hidden state at the \texttt{<|assistant|>} token, after inputting the image.
    }
    \label{fig:arch}
\end{figure*}

\clearpage
\begin{figure*}[tb]
    \centering
    \begin{subfigure}[b]{\textwidth}
         \centering
         \includegraphics[width=\textwidth]{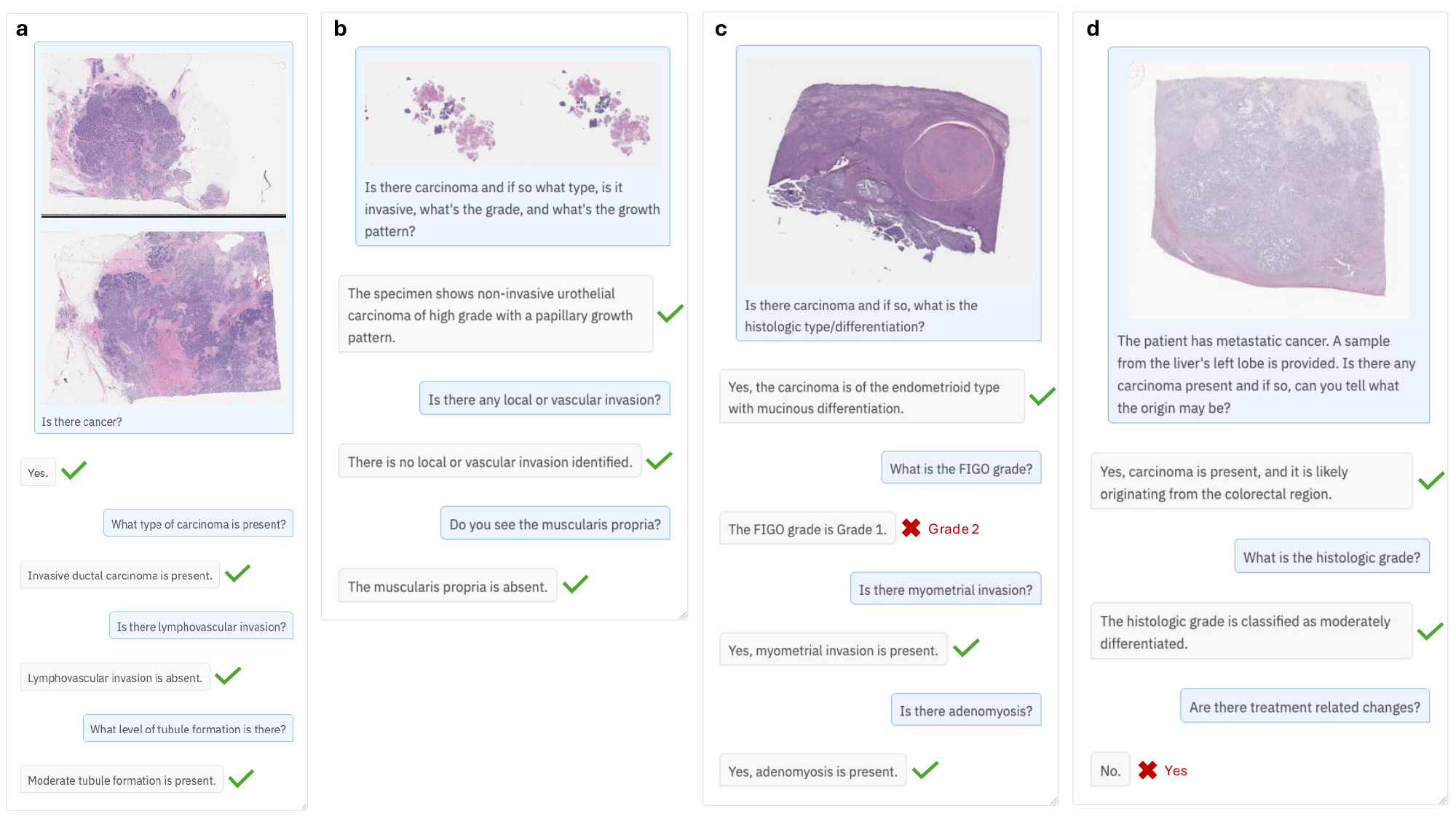}
     \end{subfigure}
    \caption{This figure demonstrates the examples of PRISM2 dialogues. \textbf{a} A breast specimen with IDC. We show \acp{WSI} in the initial message for illustrative purposes, but for each example PRISM2 views all the slides of the specimen at once. \textbf{b} A bladder biopsy. Despite being trained on simple questions, the model is able to accurately respond to compound questions. \textbf{c} A uterus specimen. The model correctly identifies and characterizes the carcinoma and additional findings except it incorrectly predicts FIGO Grade 1 instead of Grade 2. \textbf{d} A liver specimen from a patient with metastatic colorectal cancer. The model is able to be given additional context like some patient history and specimen description and detects the carcinoma and correctly predicts its origin and grade, but incorrectly says there are no treatment related changes.}
    \label{fig:extended_dialogue}
\end{figure*}

\begin{figure*}[p]
    \centering
    \begin{subfigure}[b]{\textwidth}
         \centering
         \includegraphics[width=0.9\textwidth]{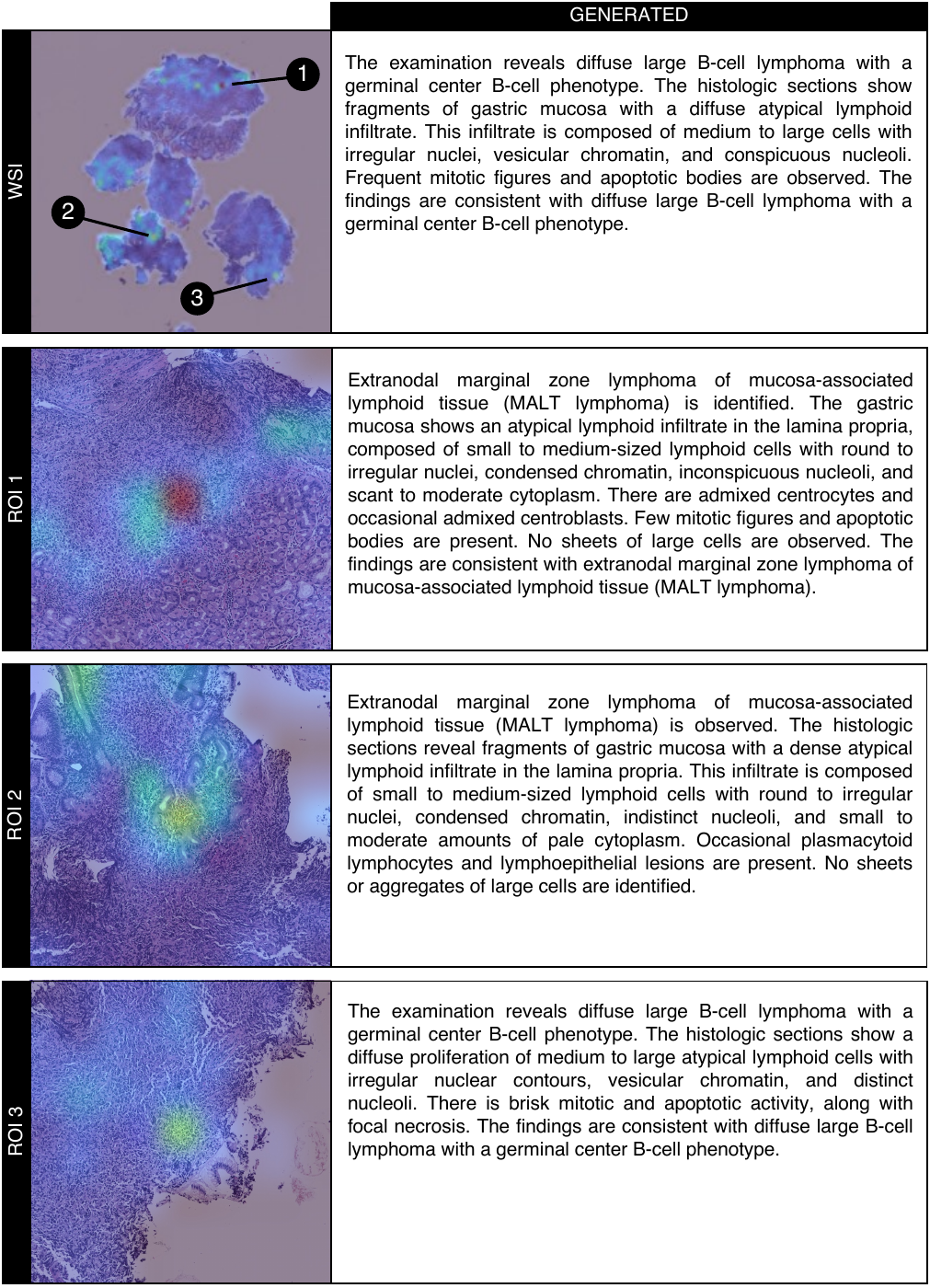}
     \end{subfigure}
    \caption{The slide encoder attention heatmap is shown for a core in a single slide of a lymphoma specimen, along with the diagnostic report generated by PRISM2. When cropping to each of the three top-attended regions, the PRISM2 report remains consistent with the evidence in the regions of interest.}
    \label{fig:extended_roi_lymphoma}
\end{figure*}

\begin{figure*}[p]
    \centering
    \begin{subfigure}[b]{\textwidth}
         \centering
         \includegraphics[width=0.9\textwidth]{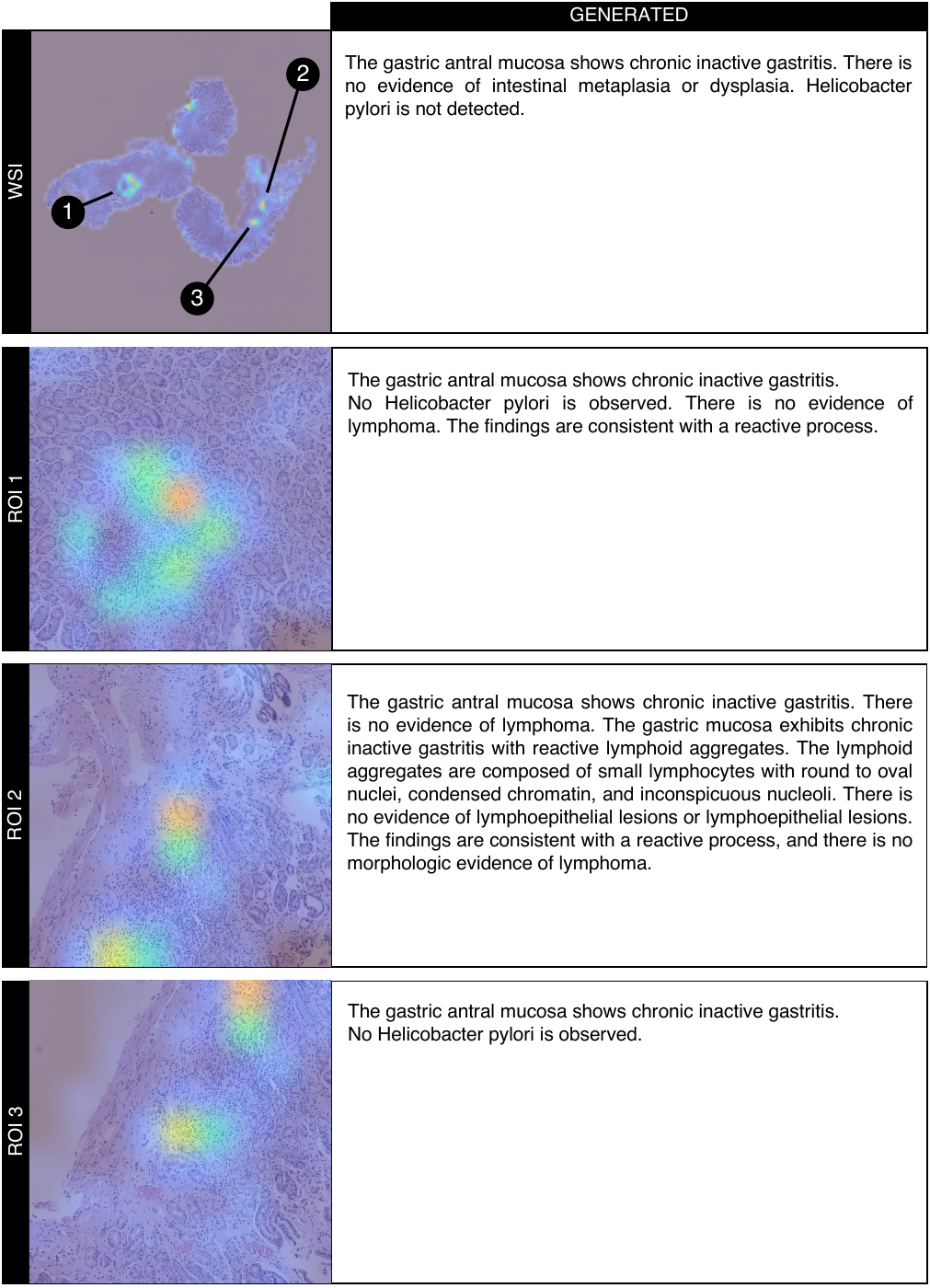}
     \end{subfigure}
    \caption{The slide encoder attention heatmap is shown for a core in a single slide of a gastric specimen, along with the diagnostic report generated by PRISM2. When cropping to each of the three top-attended regions, the PRISM2 report remains consistent with the evidence in the regions of interest.}
    \label{fig:extended_roi_gastric}
\end{figure*}

\clearpage
\begin{figure*}[hbt]
    \centering
    \begin{subfigure}[b]{\textwidth}
         \centering
         \includegraphics[width=\textwidth]{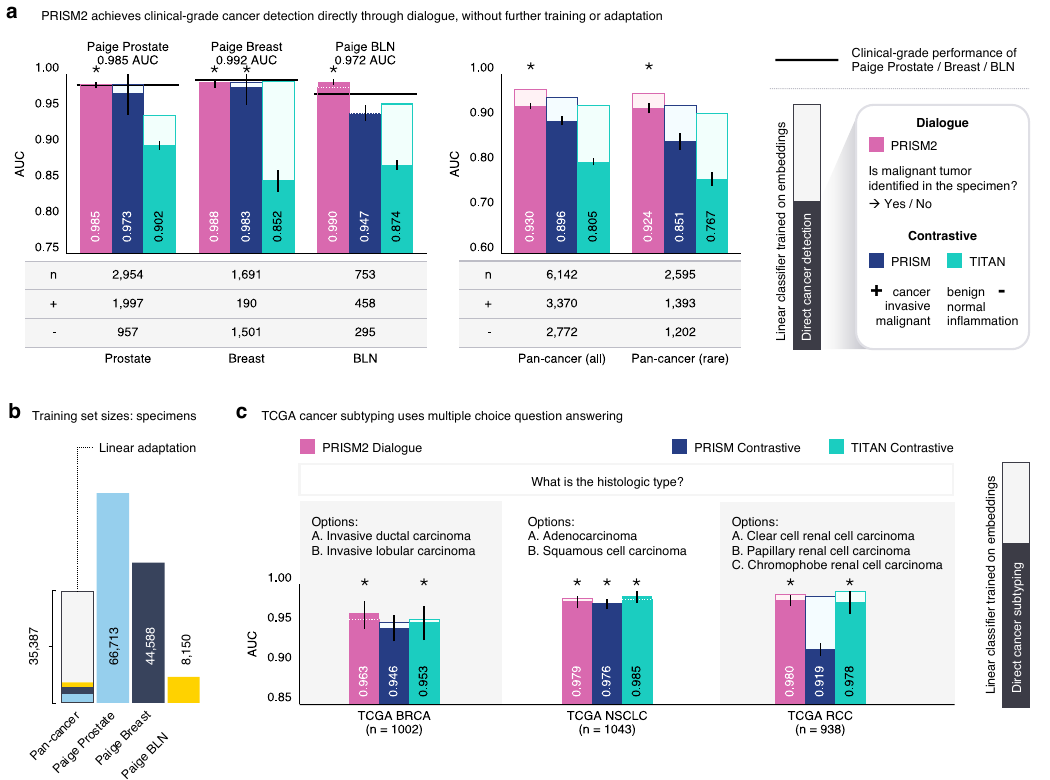}
     \end{subfigure}
    \caption{
    Direct prediction through dialogue with PRISM2 (\ac{AUC}). \textbf{a} Yes-no \ac{QA} matches or exceeds clinical-grade AI product performance for the detection of invasive cancer as compared to Paige Prostate, Paige Breast, and Paige Breast Lymph Node (BLN), when evaluated on the corresponding product testing datasets. When training a linear classifier on a large pan-cancer dataset, direct prediction by PRISM2 is as good as the linear classifier on the product testing datasets (Prostate, Breast, BLN) and nearly as good as the linear classifier on the pan-cancer dataset, including the rare cancers subset. \textbf{b} The size of the pan-cancer training set used for linear adaptation is similar to the training sets of the three clinical models; however, prostate (light blue), breast (dark blue), and BLN (yellow) tissues are a relatively small subset of the pan-cancer dataset. \textbf{c} Direct prediction performs well on \ac{TCGA} cancer subtyping with multiple-choice \ac{QA}. \textit{n} indicates number of samples; $+$ and $-$ indicate number of positive and negative samples, respectively. Error bars show the 95\% confidence interval computed with bootstrapping. * direct prediction result tied for first place (p < 0.05, permutation test).
    }
    \label{fig:extended_direct_prediction_results}
\end{figure*}

\clearpage
\begin{figure*}[hbt]
    \centering
    \begin{subfigure}[b]{\textwidth}
         \centering
         \includegraphics[width=\textwidth]{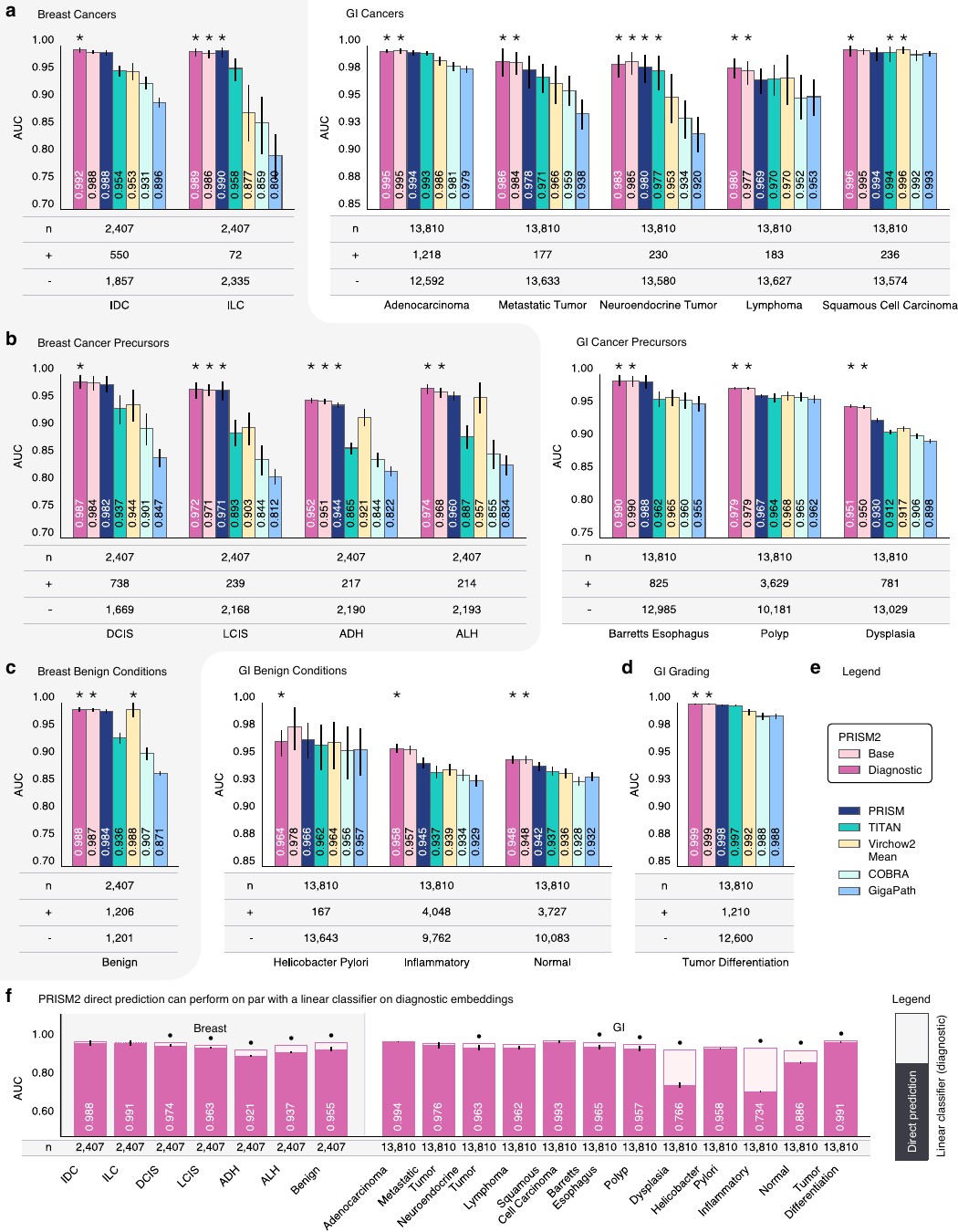}
     \end{subfigure}
     \phantomcaption
    \label{fig:extended_diagnostic_results}
\end{figure*}
\begin{figure*}[t!]
    \ContinuedFloat
    \caption{
    Adapting slide-level model embeddings to downstream diagnostic pathology tasks with a linear classifier. The detection of breast and GI cancers. Panels \textbf{a}-\textbf{d} show the \ac{AUC} performance for the detection of cancers (\textbf{a}), precursors to cancer (\textbf{b}), and benign conditions (\textbf{c}) for breast and \acf{GI} tissues, as well as cancer grading for \ac{GI} (\textbf{d}); legend in \textbf{e}. \textbf{f} Yes-no \ac{QA} enables direct detection without further training at or near the performance of the linear classifier, especially for cancers. \textit{n} indicates number of samples; $+$ and $-$ indicate number of positive and negative samples, respectively. Error bars show 95\% confidence intervals computed with bootstrapping. * linear probe result tied for first place (p < 0.05, permutation test); $\bullet$ direct prediction differs from linear probe result (p < 0.05, permutation test).
    }
\end{figure*}

\vfill

\clearpage
\begin{figure*}[hbt]
    \centering
    \begin{subfigure}[b]{\textwidth}
         \centering
         \includegraphics[width=\textwidth]{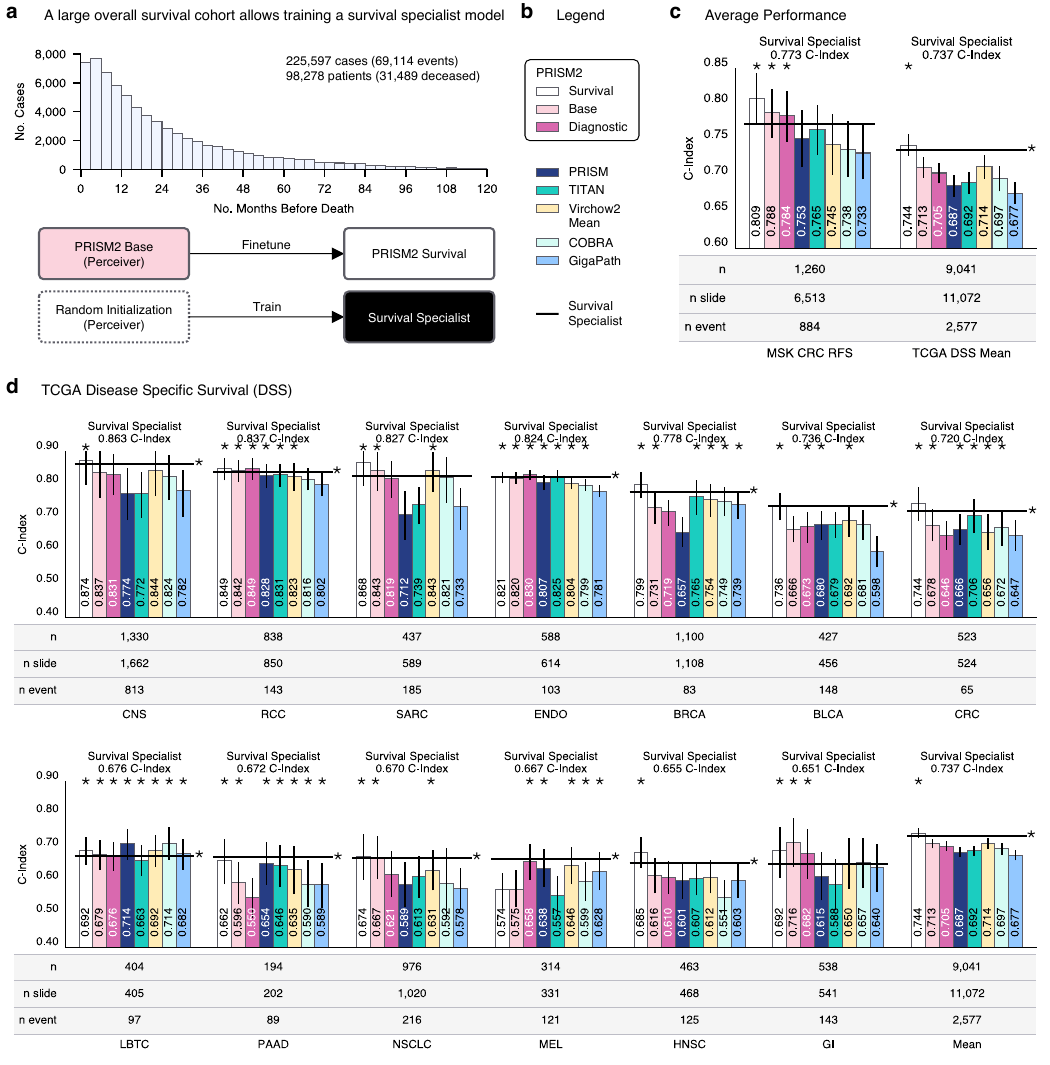}
     \end{subfigure}
    \caption{
    Adapting slide-level model embeddings to survival prediction. \textbf{a} A large \acf{OS} cohort was used to train the survival specialist model as well as the PRISM2 Survival model, by fine-tuning the slide encoder (base embedding). \textbf{b}-\textbf{d} Survival task performance evaluated with linear Cox regression (C-Index) for \Acf{CRC} \acf{RFS} with \ac{MSK} data and \acf{DSS} for \ac{TCGA} data (legend in \textbf{b}). \textbf{c} PRISM2 Survival embeddings outperform all others and match or exceed the performance of the survival specialist model, followed by PRISM2 Base. \textbf{d} Per-task \ac{DSS} prediction on \ac{TCGA} data. \textit{n} indicates number of samples; \textit{n slide} and \textit{n event} indicate number of \ac{WSI}s and recurrence or death events, respectively. Error bars show 95\% confidence intervals computed with bootstrapping. * survival result tied for first place (p < 0.05, permutation test).
    }
    \label{fig:extended_survival_results}
\end{figure*}

\clearpage
\begin{figure*}[hbt]
    \centering
    \begin{subfigure}[b]{\textwidth}
         \centering
         \includegraphics[width=\textwidth]{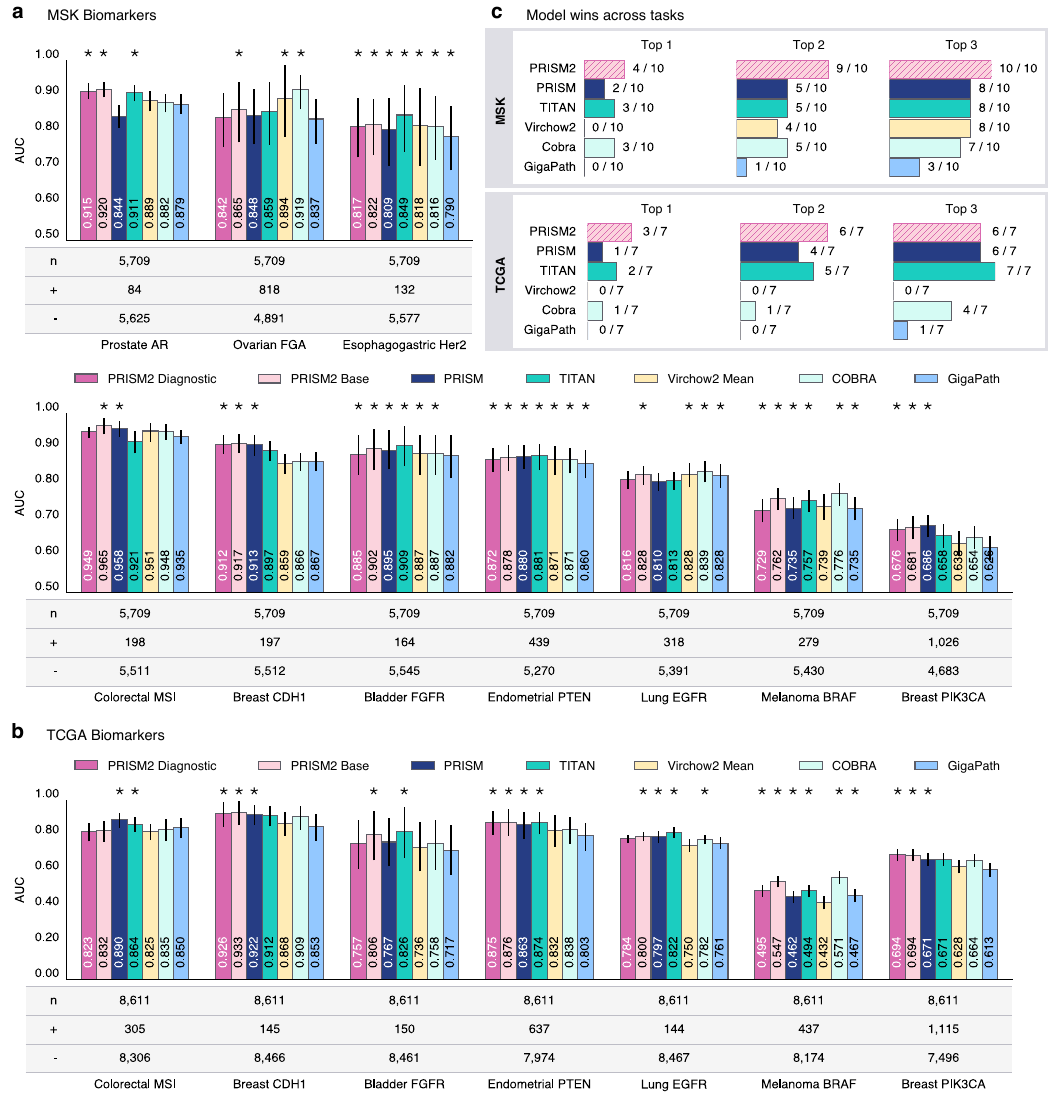}
     \end{subfigure}
    \caption{
    Adapting slide-level model embeddings to biomarker prediction. Linear classifier on model embeddings, evaluated with \ac{AUC}. \textbf{a} \ac{MSK} biomarker prediction tasks. \textbf{b} \ac{TCGA} biomarker prediction tasks. \textbf{c} The number of times each model's embeddings score within the top 1, 2, and 3 best-performing embeddings across tasks, as measured by the \ac{AUC} rounded to within two decimal points. \textit{n} indicates number of samples; $+$ and $-$ indicate number of positive and negative samples, respectively. Error bars show 95\% confidence intervals computed with bootstrapping. * linear classifier result tied for first place (p < 0.05, permutation test).
    }
    \label{fig:extended_biomarker_results}
\end{figure*}

\newpage
\appendix
\renewcommand{\thesection}{S}
\setcounter{table}{0}
\setcounter{figure}{0}
\renewcommand{\thetable}{S\arabic{table}}
\renewcommand{\thefigure}{S\arabic{figure}}
\clearpage
\ifnum\papermode=0
\setcounter{page}{1}
\fi
\section{Supplementary Notes}

\subsection{Direct dialogue completions and contrastive prompts} \label{sec:zero_shot_prompts}

\begin{table*}[htb]
  \centering
    \begin{tabular}{p{0.26\textwidth}p{0.7\textwidth}}
    \toprule
    Task & Prompt \\
    \midrule
    \multicolumn{2}{c}{Pan-cancer (all and rare), Prostate / Breast / \ac{BLN} Product Testing Dataset} \\
    \midrule
    Cancer & Is malignant tumor identified in the specimen? \{\textbf{Yes},\textbf{No}\} \\
    \midrule
    \multicolumn{2}{c}{Breast Subtyping} \\
    \midrule
    Invasive ductal carcinoma & Is invasive ductal carcinoma (IDC) identified in the specimen? \{\textbf{Yes},\textbf{No}\} \\
    Invasive lobular carcinoma & Is invasive lobular carcinoma (ILC) identified in the specimen? \{\textbf{Yes},\textbf{No}\} \\
    Ductal carcinoma in situ & Is ductal carcinoma in situ (DCIS) identified in the specimen? \{\textbf{Yes},\textbf{No}\} \\
    Lobular carcinoma in situ & Is lobular carcinoma in situ (LCIS) identified in the specimen? \{\textbf{Yes},\textbf{No}\} \\
    Atypical ductal hyperplasia & Is atypical ductal hyperplasia (ADH) identified in the specimen? \{\textbf{Yes},\textbf{No}\} \\
    Atypical lobular hyperplasia & Is atypical lobular hyperplasia (ALH) identified in the specimen? \{\textbf{Yes},\textbf{No}\} \\
    Invasive carcinoma & Is invasive carcinoma identified in the specimen? \{\textbf{Yes},\textbf{No}\} \\
    Benign or unremarkable & Is the specimen benign or unremarkable? \{\textbf{Yes},\textbf{No}\} \\
    \midrule
    \multicolumn{2}{c}{GI Suite} \\
    \midrule
    Adenocarcinoma & Is adenocarcinoma identified in the specimen? \{\textbf{Yes},\textbf{No}\} \\
    Lymphoma & Is lymphoma identified in the specimen? \{\textbf{Yes},\textbf{No}\} \\
    Metastatic Tumor & Is metastatic tumor identified in the specimen? \{\textbf{Yes},\textbf{No}\} \\
    Neuroendocrine Tumor & Is neuroendocrine tumor identified in the specimen? \{\textbf{Yes},\textbf{No}\} \\
    Squamous Cell Carcinoma & Is squamous cell carcinoma identified in the specimen? \{\textbf{Yes},\textbf{No}\} \\
    Tumor Differentiation & Is malignant tumor identified in the specimen? \{\textbf{Yes},\textbf{No}\} \\
    Dysplasia & Is dysplasia identified in the specimen? \{\textbf{Yes},\textbf{No}\} \\
    Polyp & Is polyp identified in the specimen? \{\textbf{Yes},\textbf{No}\} \\
    No Significant Abnormalities & Is there no significant abnormalities identified in the specimen? \{\textbf{Yes},\textbf{No}\} \\
    H. Pylori Helicobacter & Is h pylori helicobacter identified in the specimen? \{\textbf{Yes},\textbf{No}\} \\
    Barretts Esophagus & Is intestinal metaplasia (Barrett's esophagus) identified in the specimen? \{\textbf{Yes},\textbf{No}\} \\
    Other Inflammatory Condition & Is inflammatory condition identified in the specimen? \{\textbf{Yes},\textbf{No}\} \\
    \bottomrule
    \end{tabular}%
    \caption{
       Yes-no questions used for direct \ac{QA}-based prediction evaluation on binary classification tasks.
       For each task, the language model is prompted with a question.
       Log probabilities of two selected completions - \textbf{Yes} and \textbf{No} answers - are ranked against each other, with \textbf{Yes} acting as a positive class prediction.
    }
  \label{tab:supp_dialogue_yn_prompts}%
\end{table*}%

\begin{table*}[htb]
  \centering
    \begin{tabular}{ll}
    \toprule
    Class & Prompt \\
    \midrule
    \multicolumn{2}{c}{TCGA BRCA} \\
    \midrule
    & What is the histologic type? \\
    IDC & \textbf{A}. Invasive ductal carcinoma \\
    ILC & \textbf{B}. Invasive lobular carcinoma \\
    \midrule
    \multicolumn{2}{c}{TCGA NSCLC} \\
    \midrule
    & What is the histologic type? \\
    LUAD & \textbf{A}. Adenocarcinoma \\
    LUSC & \textbf{B}. Squamous cell carcinoma \\
    \midrule
    \multicolumn{2}{c}{TCGA RCC} \\
    & What is the histologic type? \\
    CCRCC & \textbf{A}. Clear cell renal cell carcinoma \\
    PRCC & \textbf{B}. Papillary renal cell carcinoma \\
    CHRCC & \textbf{C}. Chromophobe renal cell carcinoma \\
    \bottomrule
    \end{tabular}%
    \caption{
       Multiple-choice questions used for direct \ac{QA}-based prediction evaluation on multi-class classification tasks. For each task, the language model is prompted with a question and a list of options. Log probabilities of option letters (highlighted in bold) are ranked against each other.
    }
  \label{tab:supp_dialogue_mcq_prompts}%
\end{table*}%

\begin{table*}[htb]
  \centering
    \begin{tabular}{ll}
    \toprule
    Class & Prompt \\
    \midrule
    \multicolumn{2}{c}{Pan-cancer (all and rare)} \\
    \midrule
    \multirow{7}*{Positive} & cancer \\
    & carcinoma \\
    & adenocarcinoma \\
    & malignant \\
    & metastatic \\
    & invasive \\
    & sarcoma \\
    \midrule
    \multirow{7}*{Negative} & benign \\
    & inflammation \\
    & normal \\
    & infectious \\
    & cyst \\
    & polyp \\
    & unremarkable \\
    \midrule
    \multicolumn{2}{c}{Pan-cancer (breast subset)} \\
    \midrule
    \multirow{3}*{Positive} & invasive carcinoma is identified in the specimen \\
    & invasive ductal carcinoma (IDC) is identified in the specimen \\
    & invasive lobular carcinoma (ILC) is identified in the specimen \\
    \midrule
    \multirow{5}*{Negative} & ductal carcinoma in situ (DCIS) is identified in the specimen \\
    & lobular carcinoma in situ (LCIS) is identified in the specimen \\
    & atypical ductal hyperplasia (ADH) is identified in the specimen \\
    & atypical lobular hyperplasia (ALH) is identified in the specimen \\
    & the specimen is benign or unremarkable \\
    \bottomrule
    \end{tabular}%
    \caption{
       Positive- and negative-class prompts used for contrastive direct prediction on the Pan-cancer dataset. Contrastive prediction performance is highly sensitive to the prompts used. Using the breast-specific prompts substantially improves PRISM2 contrastive prediction performance on the breast subset.
    }
  \label{tab:supp_contrastive_pancancer_prompts}%
\end{table*}%

\clearpage
\subsection{GPT-4o rewriting and question-answer pair generation prompts} \label{sec:apx_gpt4_rewrites}

Clinical history, specimen, and report rewriting prompt:

\begin{lstlisting}[style=markdownstyle]
# Role

You are a pathology assistant whose goal is to help rewrite pathology reports in order to be used for in training a whole slide image captioning AI model.

# Instruction

You are given a pathology report describing a tissue sample and sometimes a clinical history and description of the specimen under review. You need to rewrite the report as if you're examining the sample yourself using natural language. Follow these instructions carefully and to the letter:

- Rewrite the entire report in natural language in the order of appearance (i.e. clinical history -> specimen description -> diagnosis).
- Separate the sections into their own key/value pairs: "history", "specimen", "diagnosis".
- If any section is not present, fill with null value.

## History section instruction
Use the information from the "History" section provided by the user to fill in the "history" section of your output. Follow these instructions carefully to do this:

- Never include dates and times or names of people/doctors/clinics.
- Do not mention if the patient had previous appointments.
- If age and/or gender of the patient is provided, retain that information in your rewrite.

## Specimen section instruction
Use the information from the "Specimen" section provided by the user to fill in the "specimen" section of your output. Follow these instructions to achieve that:

- Rewrite the specimen description to capture specimen type and procedure used to obtain the specimen and any other relevant information.

## Diagnosis section instruction
Use the information from the "Diagnosis" and "Addenda" sections provided by the user to fill in the "diagnosis" section of your output. You must follow these instruction with extreme attention to detail and care:

- Start with the most clinically significant finding followed by the findings of lesser significance.
- Include relevant morphological details that can be observed from a whole slide image, such as cell types, structures, and patterns if they are provided.
- Do not mention anything related to the patient's history or prior samples or material.
- Do not any specific macroscopic locations within the specimen (i.e. exclude things like right, left, apex, upper quadrant, mid to base, etc.).
- Do not mention the number of specimens examined (i.e. don't talk about number of lymph nodes inspected)
- Only talk about what could be inferred from looking at a whole slide image of the specimen.
- Exclude results or mentions of immunohistochemical (IHC) studies, molecular assays, or in situ hybridization.
- Never mention physical measurements or quantities using mm or cm or inches, etc.
- Only mention the results that are obvious from the report.
- Include relevant info from "Note" or "Addenda" that follows these instructions.
- Do not use language that references the given report.
- The language should read as if you're a pathologist inspecting the specimen and actively diagnosing the sample.

# Examples

## Example of a diagnosis rewrite that properly excludes macroscopic locations
```
User: Diagnosis: Metastatic carcinoma involving right ovary and left fallopian tube, unremarkable left ovary and right fallopian tube
Assistant: "diagnosis": "Metastatic carcinoma is observed in the ovaries and fallopian tube."
```

## Examples of properly excluding immunohistochemistry (IHC) information in diagnosis
```
User: Diagnosis: Invasive Carcinoma Ductal, NOS type. Progesterone Receptor Negative (no nuclear staining). HER2 Negative (0).
Assistant: "diagnosis": "There is invasive ductal carcinoma present."
```

## Example of a diagnosis rewrite that properly omits quantifying number of specimens examined
```
User: Diagnosis: Lymph Nodes Number with metastasis: 2
Assistant: "diagnosis": "Metastasis is present in lymph nodes."
```

## Example of a diagnosis rewrite that properly omits mention of a patient's prior samples
```
User: Diagnosis: Metastatic mammary carcinoma, morphologically similar to the patient's
prior specimen.
Assistant: "diagnosis": "Metastatic mammary carcinoma is detected."
```

## Example of a diagnosis rewrite that properly excludes measurements
```
User: Diagnosis: Keratinizing dysplasia, moderate.  Depth of invasion is 9 mm.
Assistant: "diagnosis": "Examination reveals moderate keratinizing dysplasia."
```

# Output
Generate five distinct summaries with a variety of linguistic styles with proper punctuation but still accurate to the provided information. Output the resulting 5 summaries as a list in JSON format. Don't output anything else. Do not wrap the output in markdown. The JSON structure should be formatted like:

{reports: [{"history": ..., "specimen": ..., "diagnosis":...}, ...]}
\end{lstlisting}
\clearpage

List of findings prompt:

\begin{lstlisting}[style=markdownstyle]
You are an AI pathology assistant. You are helping process reports from whole slide images.

Please extract phrases from the pathology report which refer to tissue, diagnostic findings, margins, or prior treatment visible in a histological whole slide image, or the absence of such.

Rules:
- Break down the report into multiple findings
- List the findings in multiple lines
- Each unique finding in the report has its own line
- Each line contains only one unique finding
- Findings are written in one sentence in naturally-sounding language
- Sentences are short and contain only one unique histological finding
- If there are more findings in one line, split them into multiple lines with one finding per line
- Findings don't contain numbers
- Only keep findings that are related to visible features on a histological whole slide image
- Exclude physical measurements (e.g. "...1.9 cm") or percentages (e.g. "...5%").
- Exclude macroscopic locations (e.g. "left", "right", "sigmoid").
- Exclude quantities referring to number of slides or specimens examined (e.g. "5 lymph nodes", "3 consecutive slides")
- Exclude immunohistochemical (IHC) and molecular tests.

The objective is to extract phrases which refer to things which can be located on a histological whole slide image, or confirmed not to be present.
\end{lstlisting}

Yes-No question-answer pair generation prompt:
\begin{lstlisting}[style=markdownstyle]
You are a pathology instructor at a world class medical school. Create one yes and one no question/answer pair for each line in a given report. Do not reference the given report in your wording. Imagine the student/trainee will be looking at the specimen/sample and needs to answer these questions. Use proper punctuation.

Return them individually as a JSON:

```
{
  "1": [{"question": str, "answer": str}, {"question": str, "answer": str}],
  "2": [{"question": str, "answer": str}, {"question": str, "answer": str}],
  etc.
}
```
\end{lstlisting}

Complementary question-answer pair generation prompt:

\begin{lstlisting}[style=markdownstyle]
You are given a pathology report and a series of the questions. Answer the questions with just "Yes." or "No." based on the report.

Output your results in JSON format with the original questions and group every 2 questions together:

```json
{
    "1": [{"question": str, "answer": str}, {"question": str, "answer": str}],
    etc.
}
```
\end{lstlisting}

\clearpage

Open-ended question-answer pair generation prompt:
\begin{lstlisting}[style=markdownstyle]
You are a pathology instructor at a world class medical school. Create one open-ended question/answer pair for each line in a given report, where answer is provided by the given line. The question should not be answerable with simply yes/no. Re-word the answer to introduce diverse linguistic patterns. Imagine the student/trainee will be looking at the specimen/sample and needs to answer these questions with no prior knowledge of the given report. Never use the word "noted". Use proper punctuation.

Return them individually as a JSON:

```
{
  "1": {"question": str, "answer": str},
  "2": {"question": str, "answer": str},
  etc.
}
```
\end{lstlisting}

Multiple-choice question-answer pair generation prompt:
\begin{lstlisting}[style=markdownstyle]
Given a report from routine diagnostic examination of pathology WSIs for a case, create all the multiple choice questions that follow CAP protocols as closely as possible. Exclude measurements, laterality, lymph node counts, and IHC/molecular/cytology information. Do not make unnecessary reference to the history or specimen sections.

Also classify the template that you think best matches the given report according to these categories:
- Breast DCIS, Resection
- Breast DCIS, Biopsy
- Breast Phyllodes Tumor
- Breast Invasive, Resection
- Breast Invasive, Biopsy
- Central Nervous System
- Adrenal Gland
- Appendix NET
- Colon NET
- Duodenum and Ampulla NET
- Jejunum and Ileum NET
- Pancreas (Endocrine)
- Paraganglioma Pheochromocytoma
- Pituitary Neuroendocrine (PitNET)
- Stomach NET
- Thyroid
- Ampulla of Vater
- Anus, Excision
- Anus, Resection
- Appendix
- Colon and Rectum, Biopsy
- Colon and Rectum, Resection
- Distal Extrahepatic Bile Ducts
- Esophagus
- Gallbladder
- GIST, Biopsy
- GIST, Resection
- Hepatocellular Carcinoma
- Intrahepatic Bile Ducts
- Pancreas (Exocrine)
- Perihilar Bile Ducts
- Small Intestine
- Stomach
- Kidney, Biopsy
- Kidney, Resection
- Penis
- Prostate Needle Biopsy  Case Level
- Prostate Needle Biopsy  Specimen Level
- Prostate, Resection
- Prostate TURP
- Testis Orchiectomy
- Testis Lymphadenectomy
- Ureter, Renal Pelvis, Biopsy
- Ureter, Renal Pelvis, Resection
- Urethra, Biopsy
- Urethra, Resection
- Urinary Bladder, Resection
- Urinary Bladder, Biopsy
- Endometrium Uterus
- Ovary, Fallopian Tube, or Peritoneum
- Trophoblastic Tumors
- Uterine Cervix, Excision
- Uterine Cervix, Resection
- Uterine Sarcoma
- Vagina, Biopsy
- Vagina, Resection
- Vulva
- Larynx
- Oral Cavity
- Major Salivary Glands
- Nasal Cavity and Paranasal Sinuses
- Pharynx
- Cutaneous Squamous Cell Carcinoma
- Precursor and Mature Lymphoid Malignancies
- Myeloid and Mixed / Ambiguous Lineage Neoplasms
- Plasma Cell Malignancies
- Retinoblastoma
- Uveal Melanoma
- Ewing, Resection
- Ewing, Biopsy
- Germ Cell Tumor, Resection
- Germ Cell Tumor, Biopsy
- Hepatoblastoma, Resection
- Hepatoblastoma, Biopsy
- Neuroblastoma, Resection
- Neuroblastoma, Biopsy
- Rhabdomyosarcoma, Resection
- Rhabdomyosarcoma, Biopsy
- Wilms, Resection
- Wilms, Biopsy
- Invasive Melanoma of the Skin: Biopsy
- Invasive Melanoma of the Skin: Excision, Re-Excision
- Merkel Cell Carcinoma
- Lung, Resection
- Diffuse Pleural Mesothelioma
- Thymus
- Bone, Biopsy
- Bone, Resection
- Soft Tissue, Biopsy
- Soft Tissue, Resection

If there is not a matching protocol (i.e. because the sample is benign) then classify as "N/A".

Output your response in a JSON format as shown below:

Example:

user: Endometrioid carcinoma with squamous differentiation is present.

you:
{"cap_protocol": "Endometrium Uterus",
"questions": [{"question": "What is the histologic type?",
  "options": [
    "Endometrioid carcinoma, NOS",
    "Endometrioid carcinoma with squamous differentiation",
    "Endometrioid carcinoma, villoglandular variant",
    "Endometrioid carcinoma with secretory differentiation",
    "Endometrioid carcinoma, other variant (specify): ________________",
    "Serous endometrial intraepithelial carcinoma",
    "Serous carcinoma",
    "Carcinosarcoma (malignant mixed M\"ullerian tumor)",
    "Mucinous carcinoma",
    "Clear cell carcinoma",
    "Small cell neuroendocrine carcinoma",
    "Large cell neuroendocrine carcinoma",
    "Mixed cell carcinoma (specify types and percentages): __________________________",
    "Undifferentiated carcinoma",
    "Dedifferentiated carcinoma"
  ],
  "answer": "Endometrioid carcinoma with squamous differentiation"},
  etc.
  ]
}

Make sure that the answer matches one of the options.
\end{lstlisting}

\clearpage
\subsection{Efficient Implementation} \label{sec:apx_eff_imp}

\textbf{Sequence packing.} Training efficiently on \acs{WSI}s at scale presents several challenges for maximizing hardware utilization. The number of tile embeddings extracted from \acs{WSI}s can vary dramatically between specimens, introducing highly variable sequence lengths within a training batch. A typical way to handle this is by padding to the longest sequence length in the batch, but this is not optimal as it introduces unnecessary compute and memory for the padding elements. Empirically, we find the ratio of compute/memory for padding elements to non-padding elements quickly exceeds $1.0$ even with small batch sizes, which means we begin to dedicate the majority of compute/memory to padding elements. To mitigate this we implement sequence packing for the slide encoder which eliminates any need for padding and thus incurs no unnecessary computational and memory cost.

\textbf{Dynamic batching.} In multi-GPU data parallel training, variable length sequences also create load imbalance issues. For one training step, each data parallel worker (i.e. a GPU) does a forward and backward pass on its own mini-batch and then hits a global synchronization point for gradient \texttt{allreduce}. This synchronization point requires each worker to wait for the slowest worker to complete its training step, resulting in significant idle GPU time. Without careful batching, one GPU might process significantly more embeddings than another, creating severe load imbalance. Our sequence packing implementation directly enables efficient dynamic batching: we set a fixed budget of 800 thousand tile embeddings per GPU per step, then greedily pack variable-length sequences from a queue to fill this budget. This ensures each worker has the same amount of embeddings per step which improves workload balancing and mitigates idle GPU time. The number of specimens/reports per worker per step then becomes dynamic.

\textbf{Efficient grouped query attention.} We opt to use grouped query attention (GQA)~\cite{ainslie2023gqa} when doing cross attention with tile embeddings. This leads to significant memory savings during training by projecting tile embeddings to a much smaller embedding space as keys/values for cross attention. We use the official FlashAttention \cite{dao2022flashattention, dao2023flashattention2} kernel from the \texttt{flash-attn} library which has native support for GQA rather than a naïve implementation that repeats key-value heads to match query heads which would negate the potential memory savings. The \texttt{flash-attn} kernel also supports sequence packing which allows us to avoid padding/unpadding before and after attention.

\textbf{Fully sharded data parallelism and activation checkpointing.} We use fully sharded data parallelism (FSDP) to shard model weights, gradients, and optimizer states. This reduces redundant memory and enables increasing model parameters and/or batch sizes. We noticed that, due to the relatively small batch size of text tokens, training of the decoder is bottlenecked by FSDP communications. We enabled activation checkpointing which reduces memory and allows increasing the batch size while still overlapping compute and communications.

\clearpage
\subsection{Acronyms}
\begin{acronym}[MSK-IMPACT]
    \acro{ADT}{androgen deprivation therapy}
    \acro{AI}{artificial intelligence}
    \acro{ADH}{atypical ductal hyperplasia}
    \acro{ALH}{atypical lobular hyperplasia}
    \acro{AMR}{adjusted mean recall}
    \acro{AR}{androgen receptor}
    \acro{AUC}{area under the curve}
    \acro{AUROC}[AUC]{area under (the receiver operating characteristic) curve}
    \acro{BLN}{Breast Lymph Node}
    \acro{BRAF}{B-Raf Proto-Oncogene}
    \acro{BRCA}{breast cancer}
    \acro{CAP}{College of American Pathologists}
    \acro{CDH1}{cadherin 1}
    \acro{CNN}{convolutional neural network}
    \acro{CRC}{colorectal cancer}
    \acro{CRPC}{castration resistant prostate cancer}
    \acro{DCIS}{ductal carcinoma in situ}
    \acro{dMMR}{deficient mismatch repair}
    \acro{DSS}{disease-specific survival}
    \acro{EGFR}{epidermal growth factor receptor}
    \acro{FGA}{fraction of genome altered}
    \acrodefplural{FGA}{fractions of genome altered}
    \acro{FGFR}{fibroblast growth factor receptor}
    \acro{GI}{gastrointestinal}
    \acro{GQA}{grouped query attention}
    \acro{HER2}{human epidermal growth factor receptor 2}
    \acro{HE}[H\&E]{hematoxylin and eosin}
    \acro{HGSOC}{high-grade serous ovarian cancer}
    \acro{HIPT}{the hierarchical image pyramid transformer}
    \acro{IDC}{invasive ductal carcinoma}
    \acro{IHC}{immunohistochemistry}
    \acro{ILC}{invasive lobular carcinoma}
    \acro{ITC}{infiltrating tumor cell}
    \acro{LCIS}{lobular carcinoma in situ}
    \acro{LLM}{large language model}
    \acro{LOH}{loss-of-heterozygosity}
    \acro{LUAD}{lung adenocarcinoma}
    \acro{LUSC}{lung squamous cell carcinoma}
    \acro{MIL}{multiple instance learning}
    \acro{MLP}{multi-layer perceptron}
    \acro{MMR}{mismatch repair}
    \acro{mpp}{microns-per-pixel}
    \acro{MSI-H}{high-frequency MSI}
    \acro{MSI}{microsatellite instability}
    \acro{MSK-IMPACT}{MSK-Integrated Mutation Profiling of Actionable Targets}
    \acro{MSK}{Memorial Sloan Kettering Cancer Center}
    \acro{MTC}{medullary thyroid cancer}
    \acro{NSCLC}{non-small cell lung cancer}
    \acro{OOD}{out-of-distribution}
    \acro{OS}{overall survival}
    \acro{PTEN}{phosphatase and tensin homolog}
    \acro{RET}{Ret Proto-Oncogene}
    \acro{RCC}{renal cell carcinoma}
    \acro{RFS}{recurrence-free survival}
    \acro{RNN}{recurrant neural network}
    \acro{TCGA}{The Cancer Genome Atlas}
    \acro{ViT}{vision transformer}
    \acro{WSI}{whole slide image}
    \acro{QA}{question answering}
    
\end{acronym}
\end{document}